# Planning for Contingencies: A Decision-based Approach

**Louise Pryor**  LOUISEP@AISB.ED.AC.UK
*Department of Artificial Intelligence, University of Edinburgh*
*80 South Bridge*
*Edinburgh EH1 1HN, Scotland*

**Gregg Collins**  COLLINS@ILS.NWU.EDU
*The Institute for the Learning Sciences, Northwestern University*
*1890 Maple Avenue*
*Evanston, IL 60201, USA*

## Abstract

A fundamental assumption made by classical AI planners is that there is no uncertainty in the world: the planner has full knowledge of the conditions under which the plan will be executed and the outcome of every action is fully predictable. These planners cannot therefore construct contingency plans, *i.e.*, plans in which different actions are performed in different circumstances. In this paper we discuss some issues that arise in the representation and construction of contingency plans and describe Cassandra, a partial-order contingency planner. Cassandra uses explicit decision-steps that enable the agent executing the plan to decide which plan branch to follow. The decision-steps in a plan result in subgoals to acquire knowledge, which are planned for in the same way as any other subgoals. Cassandra thus distinguishes the process of gathering information from the process of making decisions. The explicit representation of decisions in Cassandra allows a coherent approach to the problems of contingent planning, and provides a solid base for extensions such as the use of different decision-making procedures.

## 1. Introduction

Many plans that we use in our everyday lives specify ways of coping with various problems that might arise during their execution. In other words, they incorporate *contingency plans*. The contingencies involved in a plan are often made explicit when the plan is communicated to another agent, *e.g.*, "try taking Western Avenue, but if it's blocked use Ashland," or "crank the lawnmower once or twice, and if it still doesn't start jiggle the spark plug." So-called *classical planners*[1] cannot construct plans of this sort, due primarily to their reliance on three *perfect knowledge* assumptions:

1. The planner has full knowledge of the initial conditions in which the plan will be executed, *e.g.*, whether Western Avenue will be blocked;

2. All actions have fully predictable outcomes, *e.g.*, cranking the lawnmower will definitely either work or not work;

---

1. This category includes systems such as STRIPS (Fikes & Nilsson, 1971), HACKER (Sussman, 1975), NOAH (Sacerdoti, 1977) and MOLGEN (Stefik, 1981a, 1981b). Recent classical planners include TWEAK (Chapman, 1987), SNLP (McAllester & Rosenblitt, 1991) and UCPOP (Penberthy & Weld, 1992). The term is due to Wilkins (1988).

   3. All change in the world occurs through actions performed by the planner, *e.g.*, nobody else will use the car and empty its gas tank.

Under these assumptions the world is totally *predictable*; there is no need for contingency plans.

The perfect knowledge assumptions are an idealization of the planning context that is intended to simplify the planning process. They allow the development of planning algorithms that have provable properties such as completeness and correctness. Unfortunately, there are few domains in which they are realistic: mostly, the world is to some extent unpredictable. Relying on the perfect knowledge assumptions in an unpredictable world may prove cost-effective if the planner's uncertainty about the domain is small, or if the cost of recovering from a failure is low. In general, however, they may lead the planner to forgo options that would have been available had potential problems been anticipated in advance. For example, on the assumption that the weather will be sunny, as forecast, you may neglect to take along an umbrella; if the forecast later turns out to be erroneous, it is then impossible to use the umbrella to stay dry. When the cost of recovering from failure is high, failing to prepare for possible problems in advance can be an expensive mistake. In order to avoid mistakes of this sort, an autonomous agent in a complex domain must be able to make and execute contingency plans.

Recently, we and a number of other researchers have begun investigating the possibility of relaxing the perfect knowledge assumptions while staying close to the framework of classical planning (Etzioni, Hanks, Weld, Draper, Lesh, & Williamson, 1992; Peot & Smith, 1992; Pryor & Collins, 1993; Draper, Hanks, & Weld, 1994a; Goldman & Boddy, 1994a). Our work is embodied in Cassandra,[2] a contingency planner whose plans have the following features:

- They include specific decision steps to determine which of the possible courses of action to pursue;

- Information gathering steps are distinct from decision-steps;

- The circumstances in which it is *possible* to perform an action are distinguished from those in which it is *necessary* to perform it.

## 1.1 Issues for a Contingency Planner

A contingency planner must be able to construct plans that can be expected to succeed despite unknown initial conditions and uncertain outcomes of nondeterministic actions. An effective contingency planner must possess the following capabilities:

- It must be able to anticipate outcomes of nondeterministic actions;

- It must be able to recognize when an uncertain outcome threatens the achievement of a goal;

- It must be able to make contingency plans for all possible outcomes of the various sources of uncertainty that affect a given plan;

---

2. Cassandra was a Trojan prophet who was fated not to be believed when she accurately predicted future disasters. An earlier version of Cassandra was described in (Pryor & Collins, 1993).

- It must be able to schedule sensing actions that detect the occurrence of a particular contingency;

- It must produce plans that can be executed correctly regardless of which contingency arises.

The design of Cassandra addresses these issues. However, there are several issues that have not been addressed:

- We have not considered the problem of determining whether it is worth planning for a particular outcome;

- Cassandra is not a probabilistic planner: it cannot make use of any information about the likelihood or otherwise of any events;

- We have ignored the possibility of interleaving planning and execution (but see Section 7.4);

- Cassandra does not handle exogenous events;

- The version of Cassandra described here cannot solve Moore's bomb in the toilet problem (McDermott, 1987): it can only find plans that involve deciding between courses of action that will succeed in different contingencies (but see Section 6.5.5).

Cassandra assumes that all sources of uncertainty and all their possible outcomes are known, and plans for all those that affect the achievement of its goals. It is firmly in the classical planning mold: its job is to construct plans that are guaranteed to achieve its goals. It does not decide when to plan, or what to plan for. Moreover, although we believe that Cassandra is sound and complete, it is not systematic. In addition, the current implementation is too slow to be of practical use.

### 1.2 A Note on Terminology

The word *conditional* is used in a variety of senses in the literature. We avoid its use altogether, except when describing the work of other authors who use it in specialized senses: for example, the *conditional actions* and *conditioning* of Peot and Smith (1992). We use the term *contingency plan* to refer to a plan that contains actions that may or may not actually be executed, depending on the circumstances that hold at the time. We use the term *context-dependent* to refer to action effects that depend on the context in which the action is performed.

### 1.3 Outline

In this paper we present Cassandra, describe its algorithm in some detail, discuss the approach it takes to some important issues in contingency planning, and show how it handles a variety of example problems.

We start by describing the structure of Cassandra's plans. Section 2 describes how Cassandra represents actions, including those with uncertain outcomes; explains the system of labels that allows the determination of which of the alternative courses of action in a contingency plan should be pursued; and introduces the notion of explicit decision steps.

Section 3 briefly describes the basic planning algorithm in the absence of uncertainty. Section 4 explains how the algorithm is extended to handle uncertain outcomes of actions. In particular, the structure of Cassandra's decisions is considered, as are the problems of ensuring the soundness of the plan that is constructed. The resulting algorithm is described in detail and its properties are discussed in Section 5.

In Section 6 we consider some issues that arise in contingency planning. Section 7 describes related work on planning under uncertainty. Finally, Section 8 summarizes the contributions of this work and discusses its limitations.

## 2. Cassandra's Plan Representation

Cassandra's representation of contingency plans has three major components:

- An action representation that supports uncertain outcomes;
- A plan schema;
- A system of labels for keeping track of which elements of the plan are relevant in which contingencies.

These components are described in the remainder of this section.

### 2.1 Action Representation

Cassandra's action representation is a modified form of the STRIPS operator (Fikes & Nilsson, 1971). It consists of the *preconditions* for executing an action and the *effects* that may become true as a result of executing it, as in the standard STRIPS operator. The syntax is the same as that used in UCPOP (Penberthy & Weld, 1992). As in UCPOP, action effects are more complex than standard STRIPS effects: they may have an associated set of *secondary preconditions*, which govern the occurrence of that effect (Pednault, 1988, 1991). Secondary preconditions allow the representation of *context-dependent* effects of actions, *i.e.*, effects that depend upon the context in which the action is executed. The use of secondary preconditions is critical to Cassandra's ability to represent uncertain effects, and hence nondeterministic actions, as we discuss in Section 2.1.1.

Figure 1 shows a simplified operator schema for the action of making a selection from a soft-drink machine (the effects describing how the "make another selection" indicator light is turned off are omitted). The operator describes two possible effects of carrying out the action: the effect of acquiring a soda, which depends on the secondary precondition that a soda of the selected type is available; and the effect of having the "make another selection" indicator light come on, which depends on the secondary precondition that a soda of the selected type is *not* available. Both effects depend upon the preconditions that money has been entered into the machine and that the machine is plugged in.

#### 2.1.1 Representing Uncertain Effects

An *uncertain effect* in Cassandra is a context-dependent effect with an *unknown precondition*, *i.e.*, a precondition the planner can neither knowingly perceive nor deliberately affect.

```
Action:            (make-selection ?machine ?selection)

Preconditions:     (:and (money-entered ?machine)
                         (plugged-in ?machine))

Effects:           (:when   (available ?machine ?selection)           secondary precondition
                    :effect (:and (dispensed ?selection)
                                  (:not (money-entered ?machine))))
                   (:when   (:not (available ?machine ?selection))    secondary precondition
                    :effect (another-selection-indicator-on ?machine))
```

Figure 1: Simplified representation of operating a vending machine

For example, a malfunctioning soft-drink machine may operate intermittently; if the planner is aware of the intermittent functioning, but unaware of the conditions that govern this behavior, then the correct functioning of the device depends upon an unknown precondition. From the point of view of the planner, the uncertain effect is nondeterministic; the planner cannot tell in advance whether it will occur. Clearly, this definition is fundamentally subjective: another planner with better information might be able to specify precisely the conditions under which the device functions properly, for example if it knew how the internal mechanism of the machine worked. As another example, consider what happens when a coin is tossed: in principle, given perfect knowledge of all the forces and distances involved, it would be possible to predict the outcome. In practice, such knowledge is unavailable and the effect of the action is uncertain. In principle, it would be possible to specify the conditions that would lead to the coin landing tails up; in practice, these conditions are unknown.

It is interesting to note here that in some circumstances it might be possible for a planner to learn to predict outcomes that it had hitherto regarded as uncertain: for example, if it learned how the soda machine worked. "Unknown" refers only to the current situation. Our representation would facilitate such learning, which would simply involve learning new secondary preconditions rather than a whole new action representation.

Unknown preconditions play the same syntactic role as normal preconditions within the operator schema; they are represented by expressions formed using the pseudo-predicate `:unknown`. An effect that has a secondary precondition of this type will occur only in certain contexts which cannot be distinguished by the planner from the contexts in which it will not occur.

Figure 2 depicts a simplified example of an operator with an uncertain effect—it represents the action of operating a soft-drink machine that intermittently fails to dispense a soda despite being operated correctly. This operator has two uncertain effects, one in which the soda is dispensed, the other in which the soda is not dispensed.

Clearly, the uncertainty with respect to both these effects stems from a single underlying source, namely uncertainty about whether or not the machine will malfunction. In effect, the two unknown preconditions in the operator represent alternative results of this underlying source of uncertainty. This relationship is reflected in the two arguments to

```
Action:              (enter-selection ?machine)

Preconditions:       (:and (money-entered ?machine)
                           (plugged-in ?machine))

Effects:             (:when (:and (available ?machine ?selection)
                                  (:unknown ?ok T))
                      :effect (dispensed ?selection))              uncertain effect
                     (:when (:and (available ?machine ?selection)
                                  (:unknown ?ok F))
                      :effect (:not (dispensed ?selection)))       uncertain effect
                     (:when (available ?machine ?selection)
                      :effect (:not (money-entered ?machine)))
                     (:when (:not (available ?machine ?selection))
                      :effect (another-selection-indicator-on ?machine))
```

Figure 2: Operating a faulty soft-drink machine

the :unknown pseudo-predicate, the first of which designates the source of uncertainty with which it is associated, and the second of which designates the particular outcome of the uncertainty that it represents. The possible contexts are effectively partitioned into a set of equivalence classes, with each context in the same class producing the same outcome of the uncertainty. The outcome is then used to label the equivalence class. A condition of the form (:unknown ?class outcome) will be true if the actual context is in the class designated by outcome.

Notice that each instantiation of the operator will introduce a new source of uncertainty, which means that the first argument to the unknown precondition must be represented as a variable in the operator schema. Cassandra binds this variable to a unique identifier (*i.e.*, a skolem constant) when the operator is instantiated.

In Cassandra's representation it is assumed that different sources of uncertainty are independent of each other. No source of uncertainty can be linked to uncertain outcomes in more than one operator, but a single operator may introduce any number of sources of uncertainty, each of which may have any number of outcomes. Each source of uncertainty has an exhaustive set of mutually exclusive outcomes, each with a unique name.

### 2.1.2 Representing Other Sources of Uncertainty

A key element of Cassandra's design is the use of a single format to represent all sources of uncertainty that affect planning. In particular, all uncertainty is assumed to be manifest in uncertain effects of planning operators, as outlined above. Uncertainty about initial conditions can be handled within this format by treating initial conditions as though they were the effects of a phantom "start step" action. This treatment of initial conditions, which was initially developed for reasons unrelated to the problem of representing uncertain outcomes, is common to the SNLP family of planners to which Cassandra belongs.

Cassandra's formulation ignores uncertainty that might stem from outside interference during the execution of the agent's plans, except inasmuch as it can be represented as incomplete knowledge of initial conditions. This is, of course, a limitation of classical planners in general; all change in the world is assumed to be caused directly by the actions of the agent.

### 2.2 Basic Plan Representation

Cassandra's plan representation is an extension of that used in UCPOP (Penberthy & Weld, 1992) and SNLP (McAllester & Rosenblitt, 1991; Barrett, Soderland, & Weld, 1991), which is in turn derived from the representation used in NONLIN (Tate, 1977). A plan is represented as a schema with the following components:

- A set of *steps*;

- A set of anticipated *effects* of those steps;

- A set of *links* relating effects to the steps that produce and consume them (a step consumes an effect when it requires that effect to achieve one of its preconditions). Note that links in effect denote *protection intervals*, *i.e.*, intervals over which particular conditions must remain true in order for the plan to work properly.

- A set of *variable bindings* instantiating the operator schema;

- A *partial ordering* on the steps;

- A set of *open conditions*, *i.e.*, unestablished goals;

- A set of *unsafe links*, *i.e.*, links the conditions of which could be falsified by other effects in the plan.

A plan is *complete* when it contains no open conditions and no unsafe links.

### 2.3 Representing Contingencies

A contingency plan is intended to achieve its goal regardless of which of the foreseeable contingencies associated with it actually arise during execution. To construct a valid contingency plan, the planner must be able to enumerate these contingencies. The set of foreseeable contingencies can be computed from the sources of uncertainty that are associated with the plan. In effect, a contingency is one possible set of outcomes for all relevant sources of uncertainty.

#### 2.3.1 Contingency Labels

Keeping track of whether a plan achieves its goal in every contingency is a somewhat complex process. Cassandra, like CNLP, uses a system of labels to accomplish the necessary bookkeeping (Peot & Smith, 1992). Each goal, step, and effect in Cassandra's plan is labeled to indicate the contingencies in which that element participates:

- Goals are labeled to indicate the contingencies in which they must be achieved;

- Effects are labeled to indicate the contingencies in which they are expected to occur, *i.e.*, the contingencies in which the goals they satisfy arise;

- Steps are labeled to indicate the contingencies in which they must be performed, *i.e.*, the union of the contingencies in which any of their effects are expected to occur.

The preconditions of each effect become new goals, the labels of which correspond to the labels on the effect that give rise to them.

In general, it is assumed that a particular step could be executed in any contingency, albeit possibly to no purpose. However, it is sometimes necessary to rule a particular step out of a particular contingency as a means of preventing its interference with the plan for that contingency. For example, consider a plan to achieve the goal of having a coin heads up, the first action of which is to toss the coin (see Section 4.2.3 for a detailed discussion of this plan). In one contingency the coin lands heads up, and no further actions are required. In another contingency, the coin lands tails up and must be turned over in order for the goal to be achieved. It is clear, however, that the turning over action must not be performed in the first contingency: doing so would mean that the goal of having the coin heads up is not achieved. In Cassandra, ruling steps out is accomplished by associating negative labels with plan steps to indicate those contingencies in which the steps are not to be executed. Peot and Smith (1992) call this process *conditioning*.

In addition, every step that depends, directly or indirectly, on a particular outcome of a given source of uncertainty is ruled out of every contingency that involves an alternative outcome of that source of uncertainty. We discuss the reason for this restriction in more detail below.

Cassandra's labeling system thus provides very clear guidance to the agent executing the plan, which simply performs those steps whose positive labels reflect the actual circumstances that hold at execution. Steps with neither positive nor negative labels involving the current contingency will not affect the goals, but are not guaranteed to be executable. In contrast, the agent executing a plan produced by CNLP is guided by the reason labels attached to steps. In CNLP's plans, an action need only be executed if at least one of the goals represented in its reason labels is feasible. The agent must therefore have some method of deciding which of the top-level goals are feasible. We assume this can be done by comparing the context labels of each top-level goal (which are labeled because they are represented as dummy actions) with the circumstances that actually hold. Cassandra's method is thus simpler: the agent simply uses the positive labels of the plan steps instead of using the labels attached to a step to indicate those goals whose context labels must be analyzed.

The general principles of label propagation in Cassandra are:

- Positive labels, which denote that the plan element concerned contributes to goal achievement in that contingency, propagate along causal links from subgoals to the plan elements that establish them;

- Negative labels, which denote that the plan element concerned would prevent goal achievement in that contingency, propagate along causal links from effects to the plan elements that they establish.

The details are given in Section 5.1.4.

Planning for Contingencies: A Decision-based Approach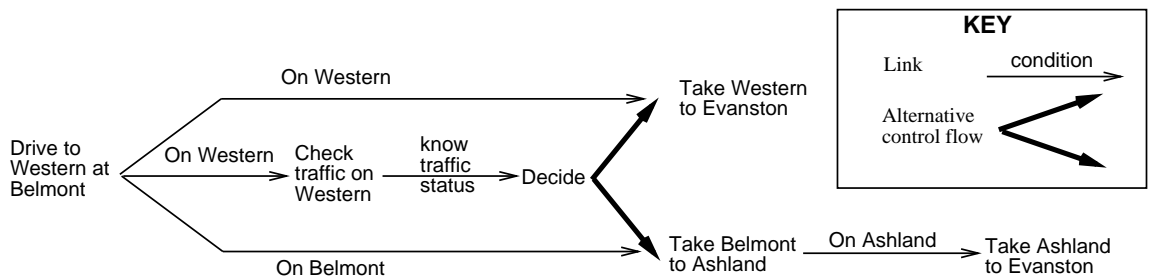

Figure 3: A plan that includes a decision-step

### 2.3.2 Representing Decisions

Planning can be seen as the process of deciding what to do in advance of when it is done (Collins, 1987). The need for contingency plans arises when the necessary decisions cannot be made in advance because of missing information (see Section 6.4). If the decisions cannot be made in advance, they must be made when the plan is executed. The agent executing a contingency plan must at some point decide which of the possible courses of action to pursue, in other words which branch to take.

Previous work has in effect assumed that the agent will execute those steps that are consistent with the contingency that actually holds (Warren, 1976; Peot & Smith, 1992). However, the determination of which steps are consistent cannot (by definition) be made in advance; in order to know which contingency holds during execution, the agent executing the plan must in general gather information on which the decision can be based. To ensure a viable plan, the planner must be able to guarantee that the steps required to gather information do not conflict with those required to carry out the rest of the plan. Therefore, the planner must in general be able to include information gathering steps, as well as any other steps that support decision making, in the plan it is constructing. Cassandra achieves this by representing decisions explicitly as plan steps. The preconditions of these decision-steps include goals to be in possession of information relevant to making the decision; the scheduling of actions to obtain information is thus handled by the normal planning process.

For instance, consider the contingency plan alluded to above: *"try taking Western Avenue, but if it's blocked use Ashland."* During the execution of such a plan, the agent must at some point decide which branch of the plan to execute. The decision-step in this case would have the precondition of knowing whether Western Avenue is blocked or not, which would cause the planner to schedule an information-gathering action to check the traffic status on Western. This operation might in turn have the precondition of being on Western, which can be achieved by traveling to the junction of Western and Belmont. After the decision is taken, the agent can either take Western up to Evanston or continue along Belmont to Ashland.

Assuming the goal of the plan is to be in Evanston, the final plan might be as depicted in Figure 3. Note that control flow after a decision is represented by heavy lines. Solid lines in the diagram represent links, with the action at the tail of the link achieving a precondition

295

of the action at the head of the link. In this plan, the agent will take Western to Evanston in one contingency, and will take Belmont to Ashland and then Ashland to Evanston in the other.[3]

Notice that in order to determine the appropriate precondition for a given decision-step, the planner must have some way of determining exactly what it will need to know in order to make the decision at execution time. This somewhat complex determination depends in part on how the decision-making process is to be carried out. In Cassandra, decisions are modeled as the evaluation of a set of condition-action rules of the form:

$$\begin{array}{lllll} \text{if} & condition_1 & \text{then} & contingency_1 \\ \text{if} & condition_2 & \text{then} & contingency_2 \\ & \ldots & & \\ \text{if} & condition_n & \text{then} & contingency_n \end{array}$$

Each possible outcome of a given uncertainty gives rise to a decision rule; the condition of this decision-rule specifies a set of effects that the agent should test in order to determine whether to execute the contingency plan for that outcome. For example, the decision-rules for the driving plan example would look like this:

$$\begin{array}{llll} \text{if} & \textit{Western Avenue is blocked} & \text{then} & \textit{execute contingency using Ashland} \\ \text{if} & \textit{Western Avenue is not blocked} & \text{then} & \textit{execute contingency using Western} \end{array}$$

Cassandra's derivation of inference rules in decisions is explained in detail in Section 4.

The preconditions for a decision-step are goals to know the truth values of the conditions in the decision-rules: they are thus *knowledge goals* (McCarthy & Hayes, 1969; Pryor, 1995) (see Section 6.4). These goals are treated in the same way as are the preconditions of any other step. Cassandra thus requires no other special provisions to allow the construction of information-gathering plans.

The explicit representation of decision-steps provides a basis for supporting alternative decision procedures. While Cassandra's basic model of the decision procedure is quite simple, more complex decision procedures can be supported within the same framework (one such procedure is described in Section 6.5.5). For example, the model could be changed to a differential-diagnosis procedure. The representation of decision procedures as templates in the same way that actions are represented as templates would allow the planner to choose between alternative methods of making a decision in the same way as it can choose between alternative methods of achieving a subgoal. An even better approach might be to formulate an explicit goal to make a correct decision, and allow the system to construct a plan to achieve that goal using inferential operators. However, this would in effect require that the goals for these operators be stated in a meta-language describing the preconditions and results of operators. We have not yet addressed this possibility in any detail.

Cassandra's separation of the gathering of information from the making of decisions allows one information-gathering step to serve several decisions. This allows the flexible use of information-gathering actions; there is no effective difference between such actions and any other action that may appear in a plan.

---

3. Appendix A shows the plans that Cassandra constructs for all the examples described in this paper. This plan is in Section A.1.

**New step** Add to the plan a new step that has an effect that will establish the open condition. Add the step preconditions and the secondary preconditions of the effect as open conditions. The open condition becomes a completed link.

**Reuse step** Make the open condition into a complete link from an effect of an existing plan step. Add the secondary preconditions of the effect as open conditions.

Figure 4: Resolving open conditions

## 3. Planning Without Contingencies

In this section we briefly review the basic planning algorithm on which Cassandra is based. It follows closely that used in UCPOP (Penberthy & Weld, 1992), which is in turn based on SNLP (McAllester & Rosenblitt, 1991). The principal difference between UCPOP and SNLP is the use of secondary preconditions (see Collins & Pryor, 1992).

Cassandra does not attempt to construct a contingency plan until it encounters an uncertainty. Up until this point, it constructs a plan in much the same manner as other planners in the SNLP family. In fact, if no uncertainty is ever introduced into the plan, Cassandra will effectively function just as UCPOP would under the same circumstances. Planning proceeds through the alternation of two processes: *resolving open conditions* and *protecting unsafe links*. Each of these processes involves a choice of methods, and may therefore give rise to several alternative ways to extend the current plan. All possible extensions are constructed, and a best-first search algorithm guides the planner's exploration of the space of partial plans.

The initial plan consists of two steps: the *start* step, with no preconditions and with the initial conditions as effects, and the *goal* step, with the goal conditions as preconditions and with no effects. The planner attempts to modify its initial plan until it is *complete*: *i.e.*, until there are no open conditions and no unsafe links.

### 3.1 Resolving Open Conditions

The planning process is driven by the need to satisfy open conditions, which are initially simply the input goals. In the course of planning to satisfy an open condition, new subgoals may be generated; these are then added to the set of open conditions. The planner can establish an open condition in one of two ways: by introducing a new step into the plan, or by reusing an existing step by making use of one of its effects (see Figure 4). The secondary preconditions of the effect that establishes the condition become open conditions. If a new step is added, the preconditions of the step become open conditions as well. Finally, each time an open condition is established, a link is added to the plan to protect the newly established condition.

One way of establishing a condition is simply to notice that the condition is true in the initial state. Because the initial conditions are treated as the results of the *start* operator, which is always a part of the plan, this method can be treated as establishment by reusing an existing step; indeed, this simplification is the motivation for representing the initial conditions in this way.

297

A link establishing the condition Cond is *unsafe* if there is an effect Eff in the plan (other than the effect SourceEff that establishes Cond and the (possible) effect GoalEff that is either established or disabled by the link) with the following properties:

**Unification**  One of the postconditions in Eff can possibly unify with either Cond or its negation;

**Ordering**  The step that produces Eff can, according to the partial order, occur both before the step that produces GoalEff and after the step that produces SourceEff.

An unsafe link may be resolved in one of three ways:

**Ordering**  Modify the ordering of the steps in the plan to ensure that the step producing Eff occurs either before the step that produces SourceEff or after the step that produces GoalEff;

**Separation**  Modify the variable bindings of the plan to ensure that the threatening effect Eff cannot in fact unify with the threatened condition Cond;

**Preservation**  Introduce a new open condition in the plan to disable Eff. This new open condition is the negation of one of Eff's secondary preconditions.

Figure 5: Unsafe links

### 3.2 Protecting Unsafe Links

Whenever an open condition is established, links in the plan may be jeopardized either because a new step threatens an existing link, or because a new link is threatened by an existing step. The situations in which a link is unsafe are shown in Figure 5. In general, a link is considered unsafe if there is an effect in the plan that could possibly interfere with the condition established by that link.

There are three general methods of protecting a threatened link (see Figure 5). First, *ordering* can be used to constrain the threatening action to occur either before the beginning or after the end of the threatened link. Second, the threatening effect and the threatened link can be *separated* by imposing constraints on the variables involved so that the effect cannot be unified with the established condition. Third, the link can be *preserved* by generating a new subgoal to disable the effect that threatens the link.

## 4. Contingency Planning

Cassandra proceeds as described in the previous section until either the plan is completed or an uncertainty is introduced. This section describes how uncertainties are introduced and how they are handled.

As an example of a plan involving an uncertainty, let us consider a version of Moore's classic "bomb in the toilet" problem (McDermott, 1987), in which the goal is *bomb is disarmed*, and the initial conditions are *bomb in package1* or *bomb in package2*. The uncertainty in this case lies in the initial conditions: depending on the outcome of the uncertainty, the *start* operator can either have the effect that the bomb is in *package1* or the effect that the bomb is in *package2*.

### 4.1 Contingencies

Uncertainty is introduced into a plan when an open condition in the plan is achieved by an uncertain effect, *i.e.*, an effect with an unknown precondition. In the bomb-in-the-toilet example, for instance, Cassandra may achieve the condition *bomb is disarmed* by selecting the dunk operator, which has the preconditions *the package is at the toilet*, and *the bomb is in the package*. The condition *the bomb is in the package* can be established by identifying it with *the bomb is in package1*, which is an effect of the *start* operator. However, this condition is uncertain, as can be determined by noting that it has an unknown precondition. Cassandra will attempt to deal with this uncertainty by introducing a new contingency (or new contingencies) into the plan. The state of the plan just after the introduction of the uncertainty is illustrated in Figure 6.

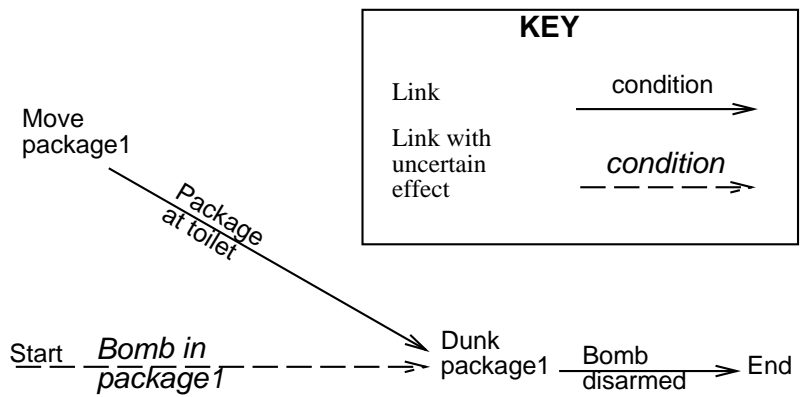

Figure 6: The introduction of uncertainty into a plan

### 4.1.1 Introducing Contingencies

Cassandra notices an uncertainty when its current plan becomes dependent upon a particular outcome of that uncertainty through the use of an uncertain effect, *i.e.*, an effect with an unknown precondition that specifies an outcome of that uncertainty. The plan that Cassandra has built up to that point is in effect a plan branch for that outcome. Since branches must also be constructed for all other possible outcomes of the uncertainty, Cassandra makes a copy of its overall goal for each possible outcome of the uncertainty, each copy carrying a label indicating the outcome of the uncertainty in which it must be achieved. It thus effectively splits the plan into a set of branches, one for each possible outcome of the uncertainty.[4]

In planning for these otherwise identical goals, Cassandra must make certain that no element of the branch for the goal for one outcome relies on a different outcome of the same uncertainty. In other words, no goal, nor any of its subgoals, may be achieved by any effect that depends, directly or indirectly, on any outcome of the uncertainty other than the one in the goal's label. As described above, Cassandra achieves this by using a system of *negative labels* indicating contingencies from which particular plan elements must be excluded.

---

4. An alternative method would be to split the plan into two branches, regardless of the number of outcomes. In this case, one branch would be associated with a given outcome of the uncertainty, while the other would be associated with all other possible outcomes of that uncertainty. This is effectively how sensp operates (Etzioni et al., 1992).

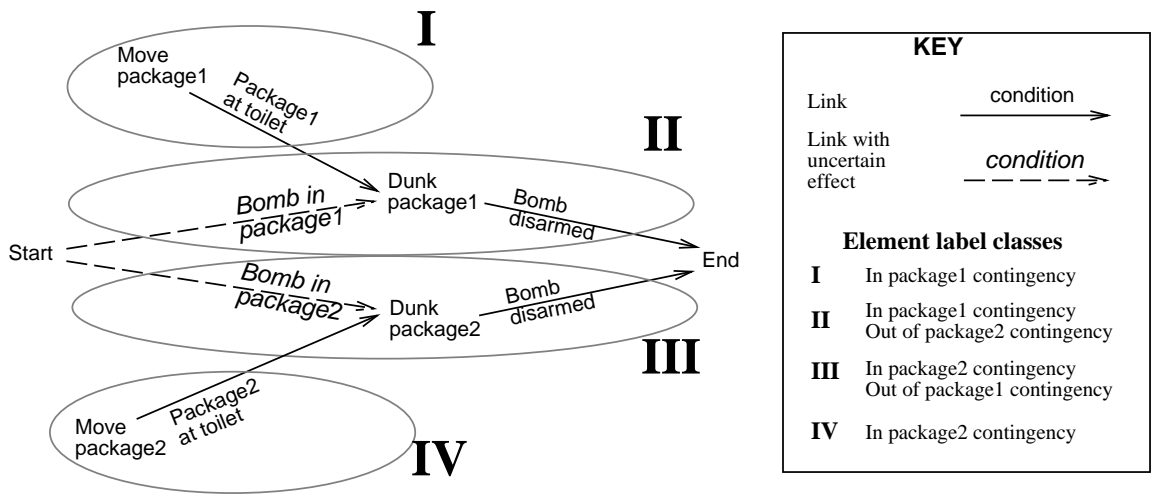

Figure 7: A contingency plan to disarm a bomb

In the bomb-in-the-toilet example, when the plan is made dependent upon the uncertain outcome *bomb in package1*, a new copy of the top level goal *bomb is disarmed* is added to the set of open conditions. The new copy is given a label indicating that it belongs to contingency in which the bomb is in *package2*.[5] The existing top level goal and all its subgoals are labeled to indicate that they belong to the contingency in which the bomb is in *package1*. The effect *bomb in package1*, the action *dunk package1*, and all effects of the action *dunk package1* are be labeled to indicate that they cannot play a role in the contingency in which the bomb is in *package2*.

Notice that the action *move package1*, although it plays a role in the plan in the contingency in which the bomb is in *package1*, does not in fact depend upon the bomb being in *package1*. It could in principle be made part of the plan for disarming the bomb in the contingency in which the bomb is in *package2*, were it to prove useful for anything. This is indicated by the fact that it has no negative label for for the *package2* contingency.

When Cassandra attempts to achieve the new open condition *bomb is disarmed*, it may choose the **dunk** operator once again (notice that it is prohibited from using any effects of the existing **dunk** operator). This new instance of the **dunk** operator in turn gives rise to a subgoal to have the bomb be in the package that is dunked. This can only be achieved by identification with the effect *bomb in package2*. The plan thus constructed is depicted in Figure 7 (the decision-step has been omitted for clarity) and is listed in Section A.2.

#### 4.1.2 Uncertainties with Multiple Outcomes

Although the algorithm we have described can deal with uncertainties having any number of possible outcomes, we have so far discussed only examples with two possible outcomes. In fact, two-outcome uncertainties suffice to describe the majority of problems that we have considered. Indeed, technically, any situation could be described in terms of some number of two-outcome uncertainties. However, it is not hard to think of situations that might naturally be represented in terms of a source of uncertainty with more than two outcomes. For example, suppose the planner were interested in getting hold of a particular object in a situation in which the object were known to be in one of three places. In such a case, the *start* pseudo-operator would naturally be represented as having three uncertain effects (one for each possible location of the object) all associated with alternative outcomes of a single source of uncertainty. Cassandra's plan for acquiring the object would then involve three contingencies, one for each possible location.

---

5. Note that we are describing the contingency in this way for clarity of exposition. The actual label is constructed as described in Section 2.3.1.

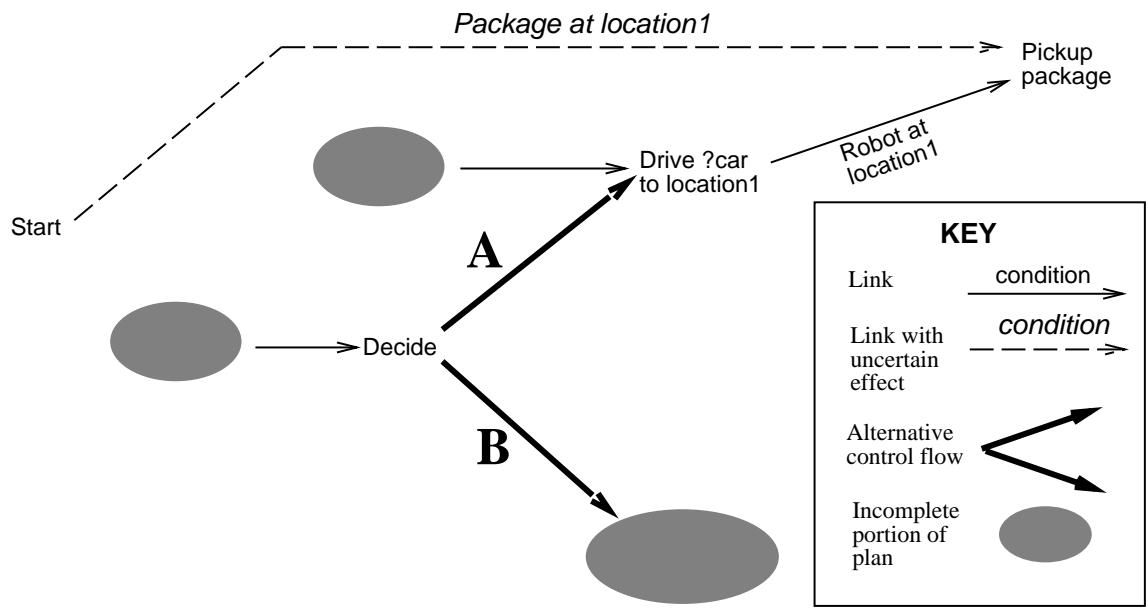

Figure 8: A partial plan to pick up a package

### 4.1.3 Multiple Sources of Uncertainty

A plan may involve two or more sources of uncertainty, in which case the plan will have more than one set of branches. For example, suppose Cassandra is given the goal of picking up a package that is at one of two locations, and that one of two cars will be available for it to use. If the uncertainty regarding the location of the package is encountered first during the construction of the plan, Cassandra will respond by building a plan involving two contingencies, one for each location. Call these contingencies $A$ and $B$ (see Figure 8 and Section A.3).

At some point during the construction of the plan for contingency $A$, Cassandra will encounter the uncertainty concerning which car will be available and will make the current plan dependent upon one particular outcome of that uncertainty. Since this new source of uncertainty arises in the context of planning for contingency $A$, contingency $A$ is in effect bifurcated into two contingencies: $A_1$, in which the package is at location 1 and car 1 is available; and $A_2$, in which the package is at location 1 and car 2 is available). Cassandra must replace all existing contingency $A$ labels with contingency $A_1$ labels. It must then introduce a new copy of the top-level goal labeled with contingency $A_2$.

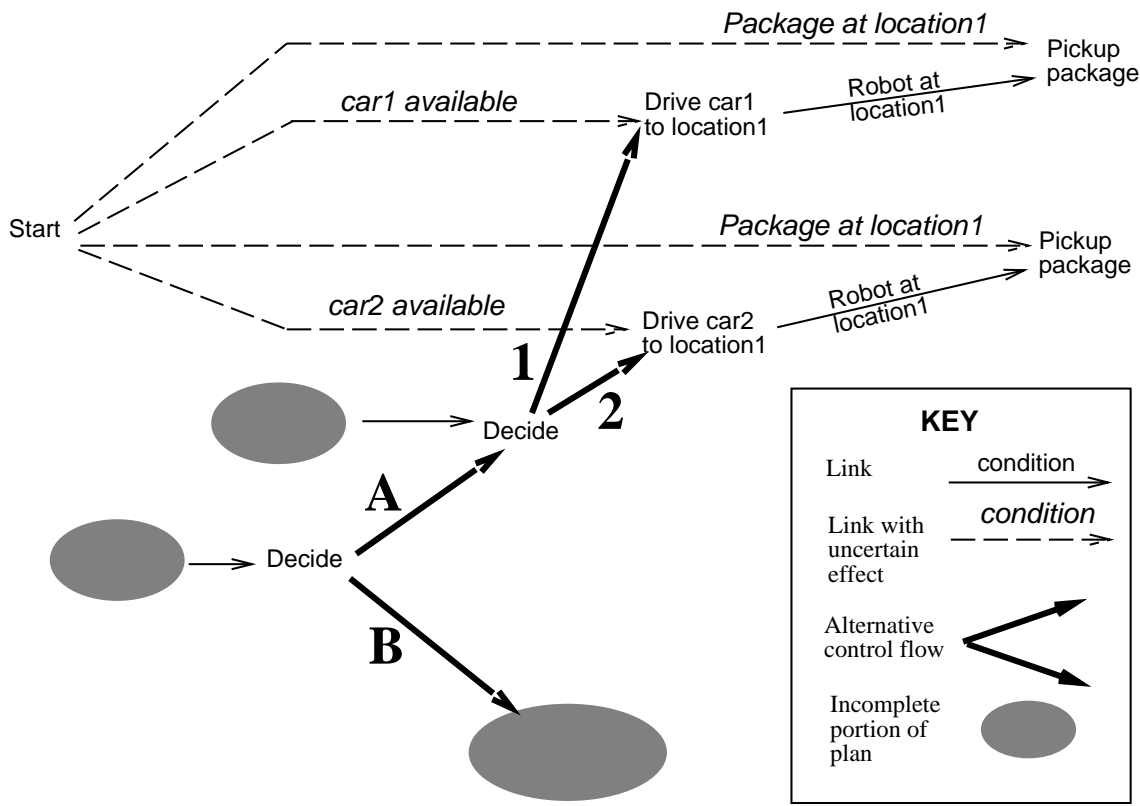

Figure 9: A plan with two sources of uncertainty

Note that Cassandra must plan from scratch to achieve the top-level goal in contingency $A_2$, in spite of the fact that it already has a viable plan for the goal in contingency $A_1$. This is necessary because situations may be encountered in which the only successful plans involve using different methods to achieve the goal in the two contingencies. For example, extreme differences between the two cars might necessitate different plans for driving them (*e.g.*, in a more detailed representation of the situation than we have presented here, such differences might affect the routes on which the cars could be driven or the places in which they could be parked). Cassandra must therefore consider all possible ways to achieve the goal in contingency $A_2$ in the search for a completion of the plan. If the particular car used does not in fact affect the driving plan, then one path through the search space will result in isomorphic contingency plans for $A_1$ and $A_2$ (see Figure 9 and Section A.4).

The same reasoning applies to the extension of the plan to deal with contingency $B$. It cannot be assumed *a priori* that the plan for contingency $B$ will in any way resemble the plan constructed for contingency $A$. An interesting consequence of this is that the uncertainty concerning the availability of the cars does not necessarily arise in a given plan for contingency $B$. For example, if the location of the package in contingency $B$ were close enough that the agent could get there without using a car, the final plan might have only three contingencies: $A_1$ (location 1 with car 1), $A_2$ (location 1 with car 2), and $B$ (location 2, on foot).

Cassandra may, of course, produce an extension of the plan in which a car is to be used in contingency $B$ as well, in which case it will again encounter the uncertainty associated with the location of the car, and will proceed to bifurcate contingency $B$ just as was done previously for contingency $A$. In the limit, the plan will involve one contingency for every member of the cross product of the possible outcomes of the relevant uncertainties. However, it is important to note that not every member of the cross-product set must appear as a contingency, since, as we have shown, some uncertainties may arise only given particular outcomes of other uncertainties.

## 4.2 Decision-steps

When Cassandra encounters a new source of uncertainty it adds a *decision-step* to the plan to represent the act of determining which path through the plan should be followed during execution. The following ordering constraints are added to the plan at the same time:

- The decision-step must occur after the step with which the uncertainty is associated;

- The decision-step must occur before any step with a precondition whose achievement depends on a particular outcome of the uncertainty.

### 4.2.1 Formulating Decision-rules

For a decision-step to be operational, there must be an effective procedure by which the agent executing the plan can determine which decision to make. In Cassandra, the action of deciding which contingency to execute is modeled as the evaluation of a set of condition-action rules of the form:

$$\begin{array}{lllll} \text{If} & condition_1 & \text{then} & contingency_1 \\ \text{If} & condition_2 & \text{then} & contingency_2 \\ \text{If} & condition_3 & \text{then} & contingency_3 \\ & \ldots & & \end{array}$$

Cassandra annotates each decision-step in a plan with the set of rules that will be used to make that decision. The executing agent can then make the decision by evaluating these rules when it comes to the decision-step in the course of executing the plan. In order to evaluate a decision-rule, the executing agent must be able to determine whether the rule's antecedent holds. The preconditions for the decision-step must thus include goals to know the current status of each condition that appears as an antecedent of a rule in this condition. The preconditions of a decision-step become open conditions in the plan in the same way as do the preconditions of any other step.

As the intended effect of evaluating the decision-rules is to choose the appropriate contingency given the outcome of a particular uncertainty, the conditions should be diagnostic of particular outcomes of the uncertainty. The executing agent cannot, of course, directly determine the outcome of an uncertainty, so it must infer it from the presence or absence of effects that depend upon that outcome.

The most straightforward approach to constructing the antecedent conditions of a decision-rule would be to analyze the plan operators to identify all the effects that could be expected to result from a given outcome of the uncertainty, and make the condition be the conjunction of these effects. However, this turns out to be overkill. In fact, it is only necessary to check for those effects of a given outcome of an uncertainty *that are actually used to establish preconditions in the contingency associated with that outcome*. In other words, it is necessary only to verify that the contingency plan can, in fact, succeed. This has the interesting consequence that the executing agent might, in principle, end up selecting a contingency plan even though the outcome of the uncertainty were not the one with which that plan was associated. Notice that this would not cause a problem in the execution of the plan, since it would only occur if all the conditions for the plan's success were met. In fact, as we shall see, Cassandra depends on this effect in certain circumstances.

The antecedent condition of the decision-rule is thus a conjunction of all the direct effects of a particular outcome that are used to establish preconditions in the contingency plan for that outcome. Decision-rules are constructed incrementally as the plan is elaborated. We discuss Cassandra's construction of these rules in more detail in Section 4.2.3 below. The approach we have used in formulating Cassandra's decision-rules is consistent with Morgenstern's observation that an agent can execute a plan if it can "make sure" that all the events in the plan are executable (Morgenstern, 1987).

### 4.2.2 Adding a Decision-rule in our Example

In the bomb-in-the-toilet example, Cassandra will introduce a decision-step to determine whether or not the bomb is in *package1*. As the uncertainty is in the initial conditions, the decision will be constrained to occur after the *start* step. It must also occur before either of the `dunk` actions, since these depend upon particular outcomes of the uncertainty. The `decide` step will have a precondition to know whether the bomb is in *package1*. If there are actions available that would allow it to determine this—X-raying the box, for example—Cassandra will achieve this precondition with one of those actions, and decide on that basis which branch of the plan to execute.

### 4.2.3 How Cassandra Constructs Decision-rules

At the point in the planning process at which Cassandra constructs a decision-rule, only one precondition in the plan is known to depend upon a particular outcome of the uncertainty that gave rise to the decision: namely, the one that led to Cassandra discovering the uncertainty in the first place. The decision-rule set that Cassandra initially builds thus looks like this:

$$
\begin{array}{llll}
\text{If} & \textit{effect}_1 & \text{then} & \textit{contingency}_1 \\
\text{If} & \text{T} & \text{then} & \textit{contingency}_2 \\
\text{If} & \text{T} & \text{then} & \textit{contingency}_3 \\
\ldots
\end{array}
$$

During the construction of the plan, Cassandra must modify this initial rule set each time an effect depending directly on the source of uncertainty is used to establish an open condition in the plan. In particular, Cassandra must determine the contingency in which that open condition resides, and conjoin the effect with the existing antecedent of the decision-rule for that contingency.

| | |
|---|---|
| Action: | `(toss-coin ?coin)` |
| Preconditions: | `(holding ?agent ?coin)` |
| Effects: | `(:when (:unknown ?U H)`  *uncertain effect* |
| |   `:effect (:and (flat ?coin)` |
| |              `(heads ?coin)))` |
| | `(:when (:unknown ?U T)` *uncertain effect* |
| |   `:effect (:and (flat ?coin)` |
| |              `(tails ?coin)))` |
| | `(:when (:unknown ?U E)` |
| |   `:effect (on-edge ?coin)))` *uncertain effect* |

Figure 10: Representing the action of tossing a coin

Consider, for example, what happens when a coin is tossed. We might say that in theory there are three possible outcomes of this action: the coin can land flat with heads up; flat with tails up; or on its edge (Figure 10). Suppose Cassandra is given a goal to have the coin be flat. This can be established by using the *flat-heads* effect of tossing it. Since this is an uncertain effect, Cassandra introduces two new contingencies into the plan, one for the outcome in which the coin lands tails up, and another for the outcome in which it lands on its edge.

The introduction of these contingencies mandates the introduction of a decision-step whose initial rule set looks like this:[6]

| | | | | |
|---|---|---|---|---|
| If | `(flat coin)` | then | `[U1: H]` | *rule for heads up* |
| If | T | then | `[U1: T]` | *rule for tails up* |
| If | T | then | `[U1: E]` | *rule for edge* |

At the same time, a new open condition `(know-if (flat coin))` is introduced as a precondition of the decision-step, and new goal conditions are introduced that must be achieved in contingencies `[U1: T]` and `[U1: E]`. Cassandra next establishes the goal condition in contingency `[U1: T]` using the *flat-tails* effect of the `toss` step. The decision-rules associated with the *tails up* contingency are thus modified as follows:

| | | | | |
|---|---|---|---|---|
| If | `(flat coin)` | then | `[U1: H]` | *rule for heads up* |
| If | `(flat coin)` | then | `[U1: T]` | *rule for tails up* |
| If | T | then | `[U1: E]` | *rule for edge* |

Finally, the goal condition is established in contingency `[U1:E]` by introducing a new step, `tip`, into the plan. A precondition of the `tip` step is that the coin be on its edge, which is established by the *on-edge* effect of the `toss` action. Since this effect depends directly

---

6. Assuming that `?U`, the variable representing the source of uncertainty, is instantiated to `U1`.

upon the uncertainty U1, the decision-rule for the *edge* contingency is modified to include this condition:

|  |  |  |  |  |
|---|---|---|---|---|
| If | (flat coin) | then | [U1: H] | *rule for heads up* |
| If | (flat coin) | then | [U1: T] | *rule for tails up* |
| If | (on-edge coin) | then | [U1: E] | *rule for edge* |

Since the plan is complete, this is the final set of decision-rules (see Section A.5). Notice that these rules do not discriminate the `heads-up` outcome from the `tails-up` outcome. In fact, either outcome will do, so there is no reason to make this discrimination. Which plan is executed in either of these conditions depends solely upon the order in which the agent that is executing the plan chooses to evaluate the decision-rules.[7]

A somewhat more complex problem arises if we give Cassandra the goal of having the coin be `flat` and `heads-up`. In this case both effects can be established using the `toss` action. This will again lead to the introduction of two new contingencies into the plan, one for when the coin lands tails up, and one for when it lands on edge. Although Cassandra could establish (flat coin) in the `tails-up` case, it would fail to complete the plan, because the coin would not be `heads-up`. However, the `turn-over` action can be used, leaving the coin `flat` and `heads-up` given that it was `flat` and `tails-up` to begin with. At this point the decision-rules are as follows:

|  |  |  |  |  |
|---|---|---|---|---|
| If | (and (flat coin) (heads-up coin)) | then | [U1: H] | *rule for heads up* |
| If | (and (flat coin) (tails-up coin)) | then | [U1: T] | *rule for tails up* |
| If | T | then | [U1: E] | *rule for edge* |

Cassandra must then plan for the goal in the outcome in which the coin lands on its edge. Both these effects can be established as a result of the `tip` action. However, the result `heads-up` is an uncertain effect of the `tip` action, since the coin might just as easily land tails up. Cassandra must therefore add another new contingency for when the coin lands tails up after being tipped. In this instance, the goal can be established by using the `turn-over` action, and the `tails-up` precondition of this action can be established by the uncertain result of the `tip` action. The final decision-rule set for the first decision is as follows:

|  |  |  |  |  |
|---|---|---|---|---|
| If | (and (flat coin) (heads-up coin)) | then | [U1: H] | *rule for heads up* |
| If | (and (flat coin) (tails-up coin)) | then | [U1: T] | *rule for tails up* |
| If | (on-edge coin) | then | [U1: E] | *rule for edge* |

If the `on-edge` contingency is pursued, another decision, stemming from the uncertain result of `tip`, must be added to the plan. If we name the second source of uncertainty U2, the rules for this decision are:

|  |  |  |  |
|---|---|---|---|
| If | (heads-up coin) | then | [U2: H] |
| If | (tails-up coin) | then | [U2: T] |

The plan is depicted in Figure 11 and shown in Section A.6.

---

7. An obvious extension to Cassandra would be the construction of a post-processor that spots decision-rules that do not discriminate between particular sets of outcomes, and prunes the plan to remove superfluous contingencies. Note that it cannot be determined until the plan is complete whether such a condition pertains.

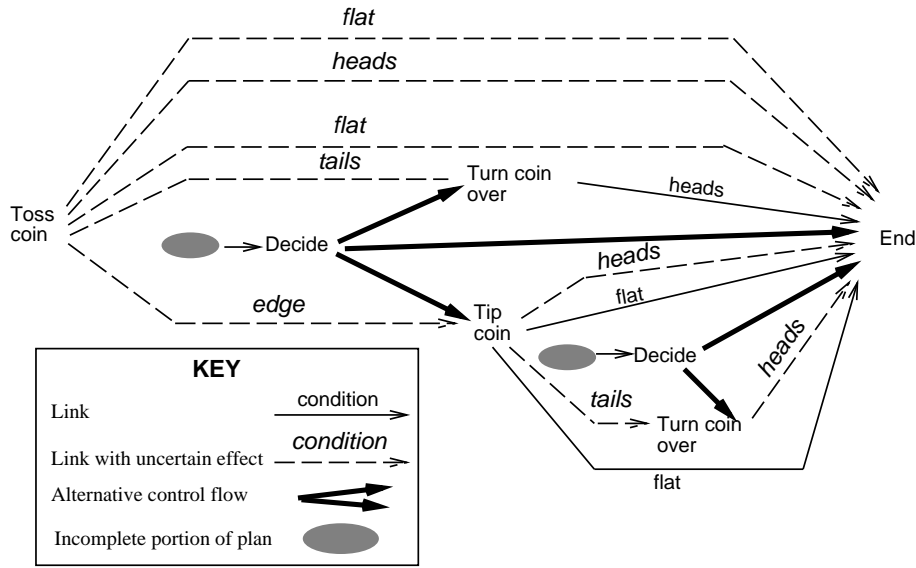

Figure 11: A plan with two decisions

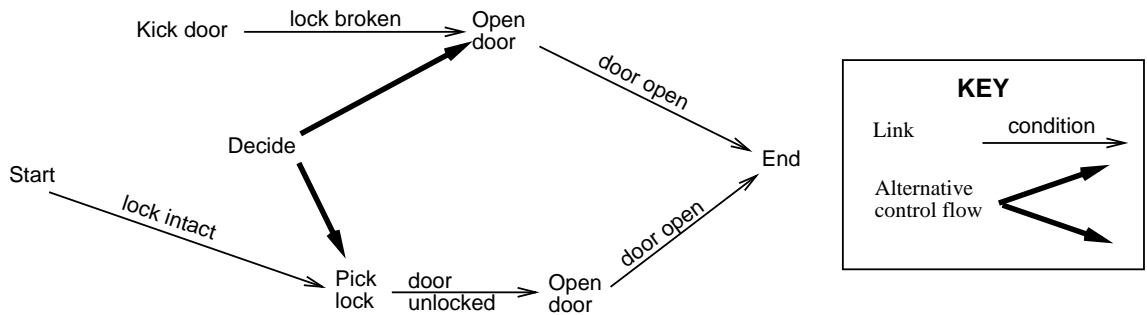

Figure 12: Opening a door

### 4.2.4 Decision-rules and Unsafe Links

The fact that Cassandra allows decision-rules that do not fully differentiate between outcomes of an uncertainty raises a somewhat subtle issue. Consider the partial plan for opening a locked door shown in Figure 12. The action of kicking a door has, let us say, two possible outcomes, one in which the lock is broken and one in which the agent's foot is broken. A plan for the contingency in which the lock is broken is simply to open the door. A plan for the alternative contingency is to pick the lock and then open the door.

Since the second plan does not depend causally on any outcome of the uncertainty (the agent's foot does not have to be broken in order for it to pick the lock and open the door), the decision-rules based on the above discussion would be:

|   | If | (lock-broken) | then | [O: L] | *rule for lock broken* |
|---|----|---------------|------|--------|-----------------------|
|   | If | T             | then | [O: F] | *rule for foot broken* |

Notice that in this case the `pick` action depends on the lock being intact, while the `kick` action may have the effect that the lock is no longer intact. In other words, the `kick` action potentially clobbers the precondition of `pick`. However, the planner can arguably ignore this clobbering, because the two actions belong to different contingencies. This is valid, though, only if the structure of the decision-rules guarantees that the agent will not choose to execute the contingency involving `pick` when the outcome of `kick` is that the lock is broken. The decision-rules above clearly do not enforce this. The solution in such a case is to augment the decision-rule for the contingency in which the lock is not broken to test whether the lock is in fact intact. This results in the following decision-rules (the plan is shown in Section A.7):

|   | If | (lock-broken)       | then | [O: L] | *rule for lock broken* |
|---|----|---------------------|------|--------|-----------------------|
|   | If | (not (lock-broken)) | then | [O: F] | *rule for foot broken* |

Cassandra augments decision-rules in this way whenever a direct effect of an uncertainty could clobber a link in a different contingency.

## 5. A Contingency Planning Algorithm

In this section we give the details of Cassandra's algorithm. Its properties are considered in Section 6.

### 5.1 Plan Elements

A plan consists of steps, effects, links (some of which may be unsafe), open conditions, variable bindings, a partial ordering, and contingency labels. A plan is *complete* when there are no open conditions and no unsafe links.

#### 5.1.1 Steps and Effects

A plan *step* Step represents an action. It may have enabling *preconditions*. It has at least one *effect* Eff. It is the instantiation of an operator.

A plan step may be a *decision-step* Decide. A decision-step has enabling preconditions of the form (know-if Cond) for a condition Cond. Decide also has a set of *decision-rules*.

An *effect* Eff represents some results of an action. It is attached to a step Step, representing that action. It may have secondary preconditions. It has at least one postcondition Cond, a condition that becomes true as the result of executing Step when the secondary preconditions hold.

#### 5.1.2 Links and Open Conditions

A *link* represents a causal dependency in the plan, specifying how a condition Cond is established by an effect Eff, which has Cond as a postcondition. Eff has secondary preconditions SecPre and is a result of step Step. The link supports the step SupStep or effect SupEff through the condition Cond which is one of:

PLANNING FOR CONTINGENCIES: A DECISION-BASED APPROACH

- An enabling precondition of `SupStep`;

- A secondary precondition of an effect `SupEff` that is a result of `SupStep`;

- The negation of a secondary precondition of an effect that is a result of `SupStep`, thus preserving a link.

A link is *unsafe* in a contingency `Conting` in which it is required if there is an effect `ClobberEff` with postcondition `ClobberCond` (the clobbering condition) resulting from step `ClobberStep` such that:

- *Either* `ClobberCond` can unify with `Cond`; 
  *Or* `Cond` is of the form (`know-if KnowCond`) and `ClobberCond` can unify with `KnowCond`;

- Step `ClobberStep` can occur between steps `Step` and `SupStep`;

- Effect `ClobberEff` can occur in contingency `Conting`.

An *open condition* (an unachieved subgoal) is represented in Cassandra as an incomplete link, *i.e.*, a link missing the information about the effect that establishes it.

### 5.1.3 BINDINGS AND ORDERINGS

Plan *bindings* (codesignation constraints) specify the relationships between variables and constants. The following relationships are possible:

- Two variables may codesignate;

- A variable may designate a constant;

- A variable may be constrained not to designate a constant;

- Two variables may constrained not to codesignate.

An *ordering* constrains the order of two steps with respect to each other, so that step $S_1$ must precede step $S_2$ ($S_1 < S_2$).

### 5.1.4 CONTINGENCY LABELS

Every step, effect and open condition in a partial plan has two sets of *contingency labels* attached to it. In the interests of brevity, we also refer to the labels of a link; in this case, we mean the labels of the step or effect that the link establishes.

Each contingency label has two parts: a symbol representing the source of uncertainty, and a symbol representing a possible outcome of that source of uncertainty. *Positive contingency labels* denote the circumstances in which a plan element must or will necessarily occur; *negative contingency labels* denote the circumstances in which a plan element cannot or must not occur.

Contingency labels must be propagated through the plan. In general, positive contingency labels are propagated from goals to the effects that establish them, while negative contingency labels are propagated from steps to the effects that result from them. The details are as follows:

*Plan*(PartList)
1. Choose a partial plan Plan from PartList;
2. If Plan is complete, then finish;
3. If there is an unsafe link Unsafe:

    Do *resolve*(Plan, Unsafe) and add the resulting plans to PartList;
    Return to step 1;

4. If there is an open condition Open:

    Do *establish*(Plan, Open) and add the resulting plans to PartList;
    Return to step 1.

Figure 13: Top level planning algorithm

- A step inherits the positive labels of the effects that result from it;

- A step inherits the negative labels of the effects that establish its enabling preconditions;

- An effect inherits the positive labels of the steps whose enabling preconditions it establishes;

- An effect inherits the positive labels of the effects whose secondary preconditions it establishes;

- An effect inherits the negative labels of the step from which it results;

- An effect inherits the negative labels of the effects that establish its secondary preconditions;

- An open condition inherits the positive labels of the step or effect that it is required to establish.

Cassandra's system of label propagation is based on that of CNLP but is more complex. Indeed, it is rather more complex than we would like. This complexity is mandated by the need to deal with operators that involve multiple context-dependent effects, which has the result that a step and its effects do not necessarily share the same labels.

### 5.2 Algorithm

The planning process starts by constructing a partial plan consisting of two steps:

- An initial step with no preconditions and with the initial conditions as its effects;

- A goal step with no effects and with the goal conditions as its enabling preconditions.

This plan is added to the (initially empty) list of partial plans PartList. Planning then proceeds as shown in Figure 13.

It now remains to describe how threats to unsafe links are resolved and how open conditions are established.

*Resolve*(`Plan`, `Unsafe`)

1. Initialize a list `NewPlans`;
2. If the unification of the clobbering condition `ClobberCond` with the condition `Cond` established by the link `Unsafe` involves adding codesignation constraints to the bindings of `Plan`:

    Make each possible modification to the bindings of `Plan` that ensures that `ClobberCond` cannot unify with `Cond`;
    Add each resulting partial plan to `NewPlans`;

3. If the clobbering step `ClobberStep` can precede the step `Step` that establishes `Unsafe`:

    Add an ordering to ensure that `ClobberStep` precedes `Step`;
    Add the resulting partial plan to `NewPlans`;

4. If the step `SupStep` supported by `Unsafe` can precede `ClobberStep`:

    Add an ordering to ensure that `SupStep` precedes `ClobberStep`;
    Add the resulting partial plan to `NewPlans`;

5. Prevent the clobbering effect `ClobberEff` from occurring in each contingency `Conting` in which the link `Unsafe` is unsafe:

    Do one of:
    (a) Add the negation of the secondary preconditions of `ClobberEff` as an open condition with positive contingency label `Conting`;
    (b) Add `Conting` to the negative contingency labels of `ClobberStep`;
    (c) Add `Conting` to the negative contingency labels of the effect `SupEff` or step `SupStep` that `Unsafe` supports;

    If appropriate modify the relevant decision-rule as discussed in Section 4.2.4;
    Add orderings to ensure that step `ClobberStep` occurs between steps `Step` and `SupStep`;
    Propagate labels as appropriate;
    Add each resulting partial plan to `NewPlans`;

6. Return `NewPlans`.

Figure 14: Resolving threats

### 5.2.1 Resolving Threats to Unsafe Links

Figure 14 shows how threats are resolved. The methods shown in steps 2, 3, and 4 are standard methods found in SNLP and UCPOP; they are often termed *separation*, *demotion*, and *promotion* respectively. We say that the methods in step 5 *disable* the threat. The methods in steps 5a and 5b ensure that the threatening effect does not occur in a given contingency. The method in step 5a is a modification of a standard method found in UCPOP and other planners that use secondary preconditions. Essentially, the idea is to prevent an effect from occurring by ensuring that the context in which it occurs cannot hold. The method in 5b prevents an effect from occurring in a contingency by forbidding the execution of the step that produces it. The method in step 5c notes that the established step or effect cannot occur in a given contingency. If any of these techniques result in inconsistent labeling of any plan element (so that, for example, it cannot occur in every contingency in which it is required) the resulting partial plan is abandoned, as it represents a dead end in the search space.

### 5.2.2 Establishing Open Conditions

Figure 15 shows the procedure used. Procedure *EstablishPre* shows the methods of adding a new step and reusing an existing step; they are essentially the methods used in UCPOP extended to reflect the need to check and propagate contingency labels.

Procedure *EstablishUnk* shows methods of adding a new decision and reusing an existing decision that are specific to Cassandra. The issues involved were discussed in Section 4.2.

## 6. Issues in Contingency Planning

Cassandra is a partial order planner directly descended from UCPOP, which is sound, complete, and systematic—all plans produced by UCPOP are guaranteed to achieve their goals, if there is a plan then UCPOP will find it, and UCPOP never revisits a partial plan. In this section we discuss these properties and related issues in the context of contingency planning.

### 6.1 Soundness

UCPOP's soundness depends on the perfect knowledge assumptions discussed in Section 1. In particular, UCPOP's plans are sound if the initial conditions are fully specified, and if all possible effects of actions are specified in the operators that represent them. If no uncertainties are involved in the plan, Cassandra is equivalent to UCPOP and therefore constructs sound plans.

If uncertainties are involved in the plan, it can no longer be assumed that the initial conditions and effects of actions are fully specified. Indeed, the uncertainties arise *because* these assumptions are violated. However, the assumptions can be adapted to account for the presence of uncertainty: it would be possible, for example, to insist that all possible initial conditions and action effects are specified. In Cassandra's representation, this means that every source of uncertainty must be specified through the use of unknown secondary preconditions, and every possible outcome of each source of uncertainty must be specified.

We conjecture that Cassandra is sound under these conditions. The proof would follow because the procedure for adding in new goals whenever a new source of uncertainty is encountered ensures that every goal is achieved in every possible outcome of the uncertainty.

### 6.2 Completeness

We conjecture that Cassandra is complete in the limited sense that, if there is a sound plan of the form that it can construct, then Cassandra will find it. We believe that this is a simple extension of UCPOP's completeness. If there are no uncertainties involved, Cassandra will always find a plan in the same way as UCPOP. The introduction of a source of uncertainty into a plan leads to the addition of new contingent goals. Cassandra will find a plan for each of these new goals in the appropriate contingency. Thus, if the goal can indeed be achieved in every contingency, Cassandra will find a plan that achieves it, as long as there is a way of determining which contingency holds.

For example, the plan to disarm a bomb that we described in Section 4.1 relies on there being a method of determining which package the bomb is in. In McDermott's presentation of this example, the two packages are indistinguishable, and the point of the example is to illustrate that there is nonetheless a plan that will succeed in disarming the bomb, namely,

*Establish*(Plan, Open)

1. If the open condition is not of type :unknown do *EstablishPre*(Plan, Open) and return the resulting list of plans;
2. If the open condition is of type :unknown with source of uncertainty Uncertainty and outcome Outcome do *EstablishUnk*(Plan, Open, Uncertainty, Outcome) and return the resulting list of plans.

*EstablishPre*(Plan, Open)

1. Initialize list NewPlans;
2. For each effect Eff resulting from a step Step in Plan
   If Eff can occur in every contingency in which Open must be established
   and if Eff can precede the step SupStep that Open is required to support
   and if there is a postcondition EffCond of Eff that can unify with condition Cond that Open is required to establish:
   - Complete the link Open by using Eff as the establishing effect;
   - Add the resulting partial plan to NewPlans;
3. For each operator with an effect Eff with a postcondition EffCond that can unify with Cond:
   - Instantiate a new step Step;
   - Complete the link Open by using Eff as the establishing effect;
   - Add the enabling preconditions of Step as open conditions;
   - Add the resulting partial plan to NewPlans;
4. For each plan in NewPlans:
   - Add an ordering to ensure that Step precedes SupStep;
   - Add the bindings necessary to ensure that EffCond unifies with Cond;
   - Add the secondary preconditions SecPre of Eff as open conditions;
   - Propagate labels as appropriate;
5. Return NewPlans.

*EstablishUnk*(Plan, Open, Uncertainty, Outcome)

1. Initialize list NewCPlans;
2. If Uncertainty is a new source of uncertainty in the plan:
   - Add a new decision-step DecStep for uncertainty Uncertainty;
   - Add new top-level goals as open conditions with the appropriate labels;
   - Add the resulting partial plan to NewCPlans;
3. If Uncertainty is an existing source of uncertainty in the plan:
   - Find an existing decision-step DecStep for uncertainty Uncertainty;
   - Add the resulting partial plan to NewCPlans;
4. For each plan in NewCPlans:
   - Modify the decision-rule in DecStep for Outcome to include Cond as an antecedent;
   - Add (know-if Cond) as an open condition required to establish DecStep;
   - Add orderings to ensure that DecStep precedes SupStep;
   - Propagate labels as appropriate;
5. Return NewCPlans.

Figure 15: Establishing open conditions

dunking both packages (McDermott, 1987). The algorithm described in the previous section cannot find a plan in this situation because it is impossible to achieve the preconditions of the decision-step that determines which package to dunk. In Section 6.5.5 we discuss this example in more detail and describe a simple extension to Cassandra that allows the correct plan (to dunk both packages) to be found.

UCPOP's completeness, like its soundness, depends on the perfect knowledge assumptions we discussed in Section 1. Cassandra's completeness depends on three extensions to these assumptions:

- All sources of uncertainty are specified;

- The specified outcomes are exhaustive;

- There are actions available that allow the determination of the outcome of any uncertainty, even if only indirectly.

Unfortunately, these conditions are necessary but not sufficient. Cassandra can only find plans if the actions that it uses to determine the contingency do not interfere with the achievement of the goal. For instance, there might be a dropping action available that would detonate any bomb inside the package that was dropped. This is certainly an action that allows the determination of the outcome of the uncertainty, but there is no sound plan that makes use of it.

In order to have a useful notion of Cassandra's completeness, we must therefore specify the form of the plans that it can construct. This problem is common to proving the completeness of any planner: for example, we do not claim that SNLP, say, is incomplete because it cannot find a plan for the bomb-in-the-toilet problem. We say instead that there is no valid plan of the form that it can construct. It is fairly simple to specify the form of the plans that SNLP can construct: they consist of partially ordered sequences of steps, all of which are to be executed. The introduction of contingencies makes the description of Cassandra's plans rather more complex; we have yet to formalize a description, but are actively working in that direction. Informally, Cassandra can only construct plans that for every source of uncertainty include a step to decide on one of the relevant plan branches. The extension of Cassandra that solves the bomb-in-the-toilet problem can do so because it can construct plans that do not meet this criterion.

### 6.3 Systematicity

UCPOP is systematic: it will never visit the same partial plan twice while searching. Cassandra, as described in this paper, is not systematic; it may visit some partial plans in the search space more than once. Consider again the plan to disarm a bomb that we discussed in Section 4.1. In this plan, there are two different ways of establishing the goal to disarm the bomb: by dunking *package1*, and by dunking *package2*. Cassandra can initially choose either way of establishing the goal, leading in each case to the introduction of a contingency and the necessity of replanning to achieve the goal in the other contingency. Both search paths arrive at the same final plan, so the search is not systematic.

Cassandra could be made systematic by insisting on handling the contingencies only in a certain order, the search path that uses the other order being treated as a dead end.

However, this extension has not been added as there is currently some debate as to the desirability of systematicity. For example, Langley (1992) argues that a non-systematic search method, iterative sampling, is often better than a systematic method, depth-first search, for problems which have multiple solutions and deep solution paths. Peot and Smith (1992) observe that the performance of a non-systematic version of SNLP was better than that of the original systematic version. They ascribed this behavior to the fact that exploring duplicate plans consumed less overhead than did ensuring systematicity.

### 6.4 Knowledge Goals

An agent executing contingency plans must be able to acquire information about the actual state of the world so that it can determine which of the possible courses of action to pursue. A system that constructs contingency plans must be able to plan for this information acquisition: in general, the acquisition process may be arbitrarily complex (Pryor & Collins, 1991).

An early and influential discussion of goals to possess knowledge about the world was that by McCarthy and Hayes (1969). Since then, various theories have been developed to account for them (*e.g.*, Moore, 1985; Haas, 1986; Morgenstern, 1987; Steel, 1995). The common thread in all this work is that knowledge goals arise from the need to specify the actions that are to be performed; in other words, from the need to make actions *operational*. Work in this area has on the whole concentrated on being able to describe and represent knowledge goals, and has largely ignored the issues involved in building planners that construct plans containing them.

The structure of Cassandra is based on the notion that knowledge goals arise out of the need to make decisions as to the actions to be performed (Pryor, 1995). In our view, planning is the process of deciding what to do in advance of when it is done (Collins, 1987). In a world conforming to the perfect knowledge assumptions of classical planning this is always possible because the world is totally predictable, and plans therefore need contain no knowledge goals. However, when those assumptions are relaxed it may not be possible to make all decisions in advance if the information necessary to make them is not available to the planner. The information may be unavailable either because of the planner's limited knowledge of the world or because the events that will nondeterministically cause the conditions that affect the decisions have not yet occurred. In both cases it may be possible for the planner to determine that a decision must be made even though it cannot at that time actually make it. In this case the planner can *defer* the decision: plan to make it in the future, when the necessary information will be available. Part of the plan is then to acquire the information; the plan thus contains knowledge goals.

Cassandra's use of "unknown" preconditions to indicate nondeterminism is thus a crucial part of its mechanism. In Cassandra, knowledge goals arise as the result of deferring decisions. These deferred decisions are represented explicitly in its plans, and themselves arise directly from the incompleteness of Cassandra's knowledge of the world, whether through the effects of nondeterministic actions or through incompletely specified initial conditions. Both these forms of uncertainty are handled in the same way: once Cassandra has recognized the need to defer a decision, the *reason* for its deferral is not important except inasmuch as it results from incomplete knowledge of the world.

The view of knowledge goals as arising from deferred decisions is basically consistent with the view that they are needed in order to make actions operational, but differs from the traditional view in that knowledge goals are not directly preconditions of physical actions, but are instead preconditions of actions that make decisions. For example, McCarthy and Hayes consider the problem of a combination safe: it is commonly held that the action of opening the safe has a precondition to know the combination. In Cassandra, however, the goal of knowing the combination would arise as a subgoal of deciding which plan branch to follow, where there would be a branch for each possible combination.[8] The branches would arise because of Cassandra's incomplete knowledge of the world: the initial conditions in which the plan will be executed are not fully specified.

Cassandra uses a variant of the syntactic approach proposed by Haas (1986) to represent knowledge goals, limiting knowledge goals to the form *know-if(fact)*. This turns out to be adequate if, as we assume, all possible outcomes of any given uncertainty are known. In general, the representation used by Cassandra, based on the STRIPS representation of add and delete lists, is less powerful than the logics proposed by either Morgenstern or Haas.

### 6.5 Miscellaneous Issues in Contingency Planning

Cassandra's approach raises a number of questions concerning the desired behavior of a contingency planner, many of which do not have obvious answers. In this section we briefly consider a few of the issues raised.

#### 6.5.1 Dependence on Outcomes and Superfluous Contingencies

The fact that a contingency plan assumes a particular outcome of an uncertainty means only that it cannot depend upon a different outcome of that uncertainty. Cassandra does not enforce any constraint that the plan must causally depend upon the outcome that it assumes. For instance, in the example described in Section 2.3.2, the plan to take Ashland does not actually depend on Western being blocked; it could be executed successfully regardless of the level of traffic on Western.

This observation raises an interesting question: If a plan for a contingency turns out not to depend on any outcome of the uncertainty that gave rise to it, would this not obviate the need for plans for alternative contingencies? For instance, in our example, it might seem sensible to execute the plan to use Ashland regardless of whether Western is blocked. It might thus seem that the planner should edit the plan in some way so as to eliminate apparently superfluous contingencies. However, it can easily be shown that a version of the plan that does not involve dependence on any outcome of the uncertainty will be generated elsewhere in the search space. In the example, this would mean that the planner would in fact consider a plan that simply involved taking Ashland. If the search heuristics penalize plans involving contingencies appropriately this other plan should be preferred to the contingency plan, all other things being equal.

---

8. This raises the obvious question as to whether planning in advance for every possibility is a sensible thing to do. See Section 7.4 for a discussion of this issue.

### 6.5.2 One-sided Contingencies

The preceding discussion notwithstanding, a plan involving no contingencies is not always superior to a plan involving a contingency. This is why a planner might in fact construct a plan like the Western/Ashland one. To take a more clear-cut example, suppose Pat needs $50 to bet on a horse. She might try to borrow the $50 from Chris, but the outcome of this action is uncertain—Chris might refuse. Alternatively, she could rob a convenience store. While the robbery plan would (we shall stipulate) involve no uncertainties, it is a bad plan for other reasons. It would be better to first try to borrow $50 from Chris, and then, if that fails, rob the convenience store. Cassandra could generate this plan. In order to make it prefer the plan to the contingency-free alternative, however, its search metric would have to take into account the estimated costs of various actions, and to perform something akin to an expected value computation. (See, for example, Feldman & Sproull, 1977; Haddawy & Hanks, 1992, for discussions of decision-theoretic measures applied to planning.) In order to execute the plan properly, it would also be necessary for it to have some way of knowing that the borrowing plan should be preferred to the robbery plan if it were possible to execute either of them.

### 6.5.3 Identical Branches

It is possible that a single plan could work just as well for several different outcomes of an uncertainty. For instance, suppose the action of asking Chris for $50 has three possible outcomes: either Pat gets the money and Chris is happy (at having had the opportunity to do a favor); or Pat gets the money and Chris is unhappy (at having been obliged to do a favor); or Pat does not get the money at all. If Pat constructs a plan in which she tries to borrow $50 from Chris to bet on a horse, then, assuming that this plan does not depend upon Chris's happiness (which it might, for example, if Pat needed to get a ride to the track from Chris), the plan will work for either the "get money + Chris happy" outcome or the "get money + Chris unhappy" outcome.

Cassandra could find such a plan, but would in effect have to find it twice—once for each outcome of the uncertainty—and it would still require a decision-step to discriminate between those outcomes. This is inefficient in two ways: the extra search time required to find what is essentially the same plan twice is wasted, and effort is put into making an unnecessary decision. We are looking into ways to avoid the former problem. The latter could be solved by a post-processor that would "merge" identical contingency plans, but we have not implemented this technique.

### 6.5.4 Branch Merging

It is possible to construct a plan in which branches split and then reunite. For instance, consider the Western/Ashland plan once again. The context in which the goal to get to Evanston arises might be an obligation to deliver a toast at a dinner to be held in an Evanston restaurant. The contingency due to uncertainty about traffic on Western Avenue would in this case seem to affect only the portion of the plan concerned with getting to Evanston; it probably has little bearing on the wording of the toast, the choice of wine, and so on. The most natural way to frame this plan might thus be to assume that regardless of which contingency is carried out, the planner will eventually arrive at a certain location in Evanston, and from that point a single plan will be developed to achieve the final goal.

Constructing the plan in this way would result in a more compact plan description, and might thus reduce the effort needed to construct the plan by avoiding, for example, the construction of multiple copies of the same subplan. We are considering methods by which branch re-merging might be achieved, but all the methods we have considered so far seem to complicate the planning process considerably.

### 6.5.5 Fail-safe Planning

As we discussed in Section 6.2, Cassandra's operation relies on being able to determine, even if only indirectly, the outcome of any uncertainty. However, this may not always be possible, and it is not a necessary precondition for the existence of a viable plan. In the bomb-in-the-toilet problem, for example, there is a valid plan that Cassandra cannot find: to dunk both packages.

This suggests a method for constructing plans in the face of uncertainty when the outcome of the uncertainty cannot be determined—what one might call *fail-safe* plans. Whenever uncertainty arises it is in principle possible that there might be a non-contingent plan that would achieve the goal whatever the outcome of the uncertainty. To find such a plan, the planner must construct a version of the contingency plan in which all actions in the contingency branches arising from the uncertainty will be executed unconditionally. Cassandra has been extended in just such a way, by adding a new type of decision, one to execute all branches in parallel (Collins & Pryor, 1995). A plan containing such a decision is only sound if none of the actions that must be performed to achieve the goal in one contingency interfere with any of the actions that must be performed in any other contingency, and the ability to perform the actions is independent of the outcome of the uncertainty. These conditions clearly hold for the bomb-in-the-toilet problem.

Cassandra can reason about this possibility because its labeling scheme distinguishes those actions that must not be performed in a given contingency from those that need not be performed. It is possible to execute all branches only if the actions in each branch may be performed (but need not) in all the other branches.

When a parallel decision is added to the plan in the extended version of Cassandra, new goals are added in the usual way but the labeling is handled differently. The branches are not separated, so that Cassandra can no longer reason that the causal links in one branch will not be affected by actions in another branch.

### 6.5.6 Contingent Failure

Cassandra can produce a plan only if it is possible to achieve the goal of the plan in all possible contingencies. Often, however, the goal cannot in fact be achieved in some outcome of the underlying uncertainty. Consider, for instance, Peot and Smith's example of trying to get to a ski resort by car, when the only road leading to the resort is either clear or blocked by snowdrifts (Peot & Smith, 1992). If the road is clear, then the goal can be achieved, but if it is blocked, all plans are doomed to failure.

No planner can be expected to recognize the impossibility of achieving a goal in the general case (Chapman, 1987). However, a possible approach is suggested by Peot and Smith. We could introduce an alternative method of resolving open goal conditions: simply assume that the goal in question fails.

This is an undesirable method of resolving open goal conditions if the subgoal is in fact achievable, so in theory plans involving contingent failure should be considered only after the planner has failed to find a plan in which all goals are achieved. This is sometimes possible, but in general the problem of determining whether there is a successful plan is undecidable. There may always be partial plans that do not involve goal failure but that cannot be completed. For example, as a partial plan is modified it may become more and more complex, the resolution of each open condition involving the introduction of more unachieved subgoals. In this case, plans involving contingent failure will never be considered unless they are ranked above some plans that do not involve contingent failure. In order to be generally useful, the approach must be weakened: instead of considering goal failure only after all other avenues of attack have failed, apply a high fixed penalty to plans involving failed goals. The aim would be to fix the penalty high enough that contingent failure would only apply in genuine cases of goals being unachievable. However, this would of necessity be a heuristic approach and completeness would be lost.

## 7. Related Work

Cassandra is constructed using UCPOP (Penberthy & Weld, 1992) as a platform. UCPOP is a partial order planner that handles actions with context-dependent effects and universally quantified preconditions and effects. UCPOP is an extension of SNLP (Barrett et al., 1991; McAllester & Rosenblitt, 1991) that uses a subset of Pednault's ADL representation (Pednault, 1989).

An early contingency planner was Warren's WARPLAN-C (1976). Contingency planning was more or less abandoned between the mid seventies and the early nineties,[9] until SENSp (Etzioni et al., 1992) and CNLP (Peot & Smith, 1992). Both SENSp and CNLP are members of the SNLP family: SENSp is, like Cassandra, based on UCPOP, and CNLP is based directly on SNLP. C-BURIDAN (Draper et al., 1994a; Draper, Hanks, & Weld, 1994b), a probabilistic contingency planner, is based on the probabilistic planner BURIDAN (Kushmerick, Hanks, & Weld, 1995) (which is itself based on SNLP) and on CNLP. PLINTH (Goldman & Boddy, 1994a, 1994b) is a total-order planner based on McDermott's PEDESTAL (1991), and is strongly influenced by CNLP in its treatment of contingency plans.

WARPLAN-C, unlike the other planners considered here, did not use a STRIPS-based action representation, but was based on predicate calculus. It could handle actions that had just two possible outcomes, and did not merge the resulting plan branches.

SENSp also differs from the other planners considered here. It represents uncertainty through the use of *run-time* variables, distinguished from ordinary variables by being treated as constants whose values are not yet known. In SENSp plan branches arise from the introduction of information-gathering steps that bind the run-time variables. SENSp handles plan branching by constructing separate plans that each achieve the goal in a particular contingency. It then combines the separate plans at a later stage, keeping the branches totally separate. SENSp thus considers contingency branches separately, rather than in parallel. Actions that achieve knowledge goals may not have preconditions in sensp: this restriction is required in order to maintain completeness.

---

9. Neither NOAH (Sacerdoti, 1977) nor Interplan (Tate, 1975) explicitly addressed issues of uncertainty, although both tackled problems involving it (Collins & Pryor, 1995).

Not surprisingly, Cassandra, CNLP, and C-BURIDAN, and to a lesser extent PLINTH, are in many respects very similar. All except PLINTH use the basic SNLP algorithm, and all use extended STRIPS representations. Cassandra differs from CNLP and PLINTH principally in the way that uncertainty is represented (Section 7.1); this difference has important implications for the handling of knowledge goals (Section 7.2). The principal difference between Cassandra and C-BURIDAN lies in the latter's use of probabilities (Section 7.3).

Contingency planning is only one approach to the problem of planning under uncertainty. The aim of contingency planning is to construct a single plan that will succeed in all circumstances: it is essentially an extension of classical planning. There are other approaches to planning under uncertainty that do not share this aim: probabilistic planners aim to construct plans that have a high probability of succeeding (Section 7.3); systems that interleave planning and execution do not attempt to plan fully in advance (Section 7.4). In both these approaches it is possible to address the problem of determining which contingencies should be planned for, which is not currently possible in Cassandra. A third approach is that of reactive planning, in which behavior is controlled by a set of reaction rules (Section 7.5).

## 7.1 The Representation of Uncertainty

In CNLP and PLINTH, uncertainty is represented through a combination of uncertain outcomes of nondeterministic actions and the effects of observing those outcomes. A three-valued logic is used: a postcondition of an action may be *true*, *false*, or *unknown*. For example, the action of tossing a coin might have the postcondition `unk(side-up ?x)`. Special *conditional actions*, each of which has an unknown precondition and several mutually exclusive sets of postconditions, are then used to observe the results of the nondeterministic actions. In the example, the operator to observe the results of tossing a coin might have the precondition `unk(side-up ?x)` with three possible outcomes: `(side-up heads)`, `(side-up tails)`, and `(side-up edge)`.

CNLP thus spreads the representation of uncertainty over both the action whose execution produces the uncertainty and the action that observes the result. A consequence of this is that CNLP cannot use the same observation action to observe the results of different actions. For example, it would require different actions to observe the results of tossing a coin (which has three possible outcomes) and tipping a coin that had landed on its edge (which has two possible outcomes).

In PLINTH, the notion of a conditional action is extended to cover any action (not only observation actions) that has nondeterministic effects *on the planner's world model*. For example, in an image-processing domain an operator to remove noise from an image may or may not succeed. However, its outcome is evident as soon as it has been applied, and no special observation action is required.

In CNLP and PLINTH, information-gathering actions are included in a plan whenever an action with uncertain effects occurs. This is necessary because the uncertainty is actually represented in the information-gathering action rather than in the action that actually produces the uncertainty. Knowledge goals are thus not represented explicitly in these two systems.

The representation used in CNLP and PLINTH arises out of the desire to use a "single model of the world, representing the planner's state of knowledge, rather than a more complex formalization including both epistemic and ground formulas" (Goldman & Boddy, 1994b). An operator therefore represents only the effects that the execution of the underlying action has on the planner's knowledge of the world, and not the effects that it has on the actual state of the world. It is, of course, important to represent how actions affect the planner's world model, but we believe that it is also important to represent how they affect the world. After all, the purpose of reasoning about actions is to achieve goals in the world, not just in the planner's world model. In particular, after the execution of a nondeterministic action its actual effects, although they may indeed be unknown to the planner, have occurred and cannot now be altered. Cassandra's representation reflects this: indeed, Cassandra can reason about the possible effects without scheduling observation actions. This means that an extension of Cassandra can, for example, solve the original bomb-in-the-toilet problem, in which there are no possible actions that will resolve the uncertainty as to which package contains the bomb: the bomb's state is not represented in the planner's world model at any stage between the beginning, when it is known to be armed, and the end, when both packages have been dunked and it is known to be safe.

A further implication of this method of representing uncertainty is the difficulty of representing actions whose uncertain effects cannot be determined through the execution of a single action. Consider, for example, a malfunctioning soda machine that has one indicator that lights when it cannot make change, and another that lights when it has run out of the product requested. Suppoe that, when it is functioning correctly, these two indicators will not light simultaneously. If it malfunctions, it must be kicked to make it work. Observing either light on its own is not enough to determine which uncertain effect (working properly or malfunctioning) has occurred.

### 7.2 Knowledge Goals

The method of representing uncertainty in CNLP and PLINTH has important implications for how knowledge goals are handled in their plans.

The acquisition of information is a planning task like any other (Pryor & Collins, 1991, 1992; Pryor, 1994). In general, the sequence of actions required to achieve a given knowledge goal may be arbitrarily complex. For example, an action to observe a tossed coin might require that the observer is in the appropriate location; in other cases, there might be several different possible methods of information gathering, some involving perception, some involving reasoning, and some a combination. A contingency planner, some of whose plans will necessarily involve the achievement of knowledge goals, must therefore be able to plan fully generally for information gathering.

The confusion between the source of uncertainty and the observation of uncertain results limits the ways in which knowledge goals can be achieved in CNLP and PLINTH: they must be achieved through the special observation actions that specify the uncertain outcomes. This is a result of their representation in terms of the planner's world model, which means that they do not represent the effects of actions (except to flag them as unknown) until the planner has observed them (or otherwise incorporated them into its world model). In their discussion of this issue Goldman and Boddy (1994b) explicitly exclude knowledge goals from consideration. As they point out, planning under uncertainty requires that a distinction be made between the actual state of the world and the planner's knowledge of it. In order to plan effectively for knowledge goals, both must be represented. This is done in Cassandra by separating the representation of uncertainty from the representation of information-gathering. If an effect results deterministically from an action, Cassandra reasons that there is no need to observe it, and it forms part of the world model. An uncertain effect, on the other hand, is not incorporated unconditionally into Cassandra's world model; it is noted as being possibly true, and (if necessary) Cassandra sets up a subgoal to determine whether it is indeed true.

SENSp, which uses the UWL representation for goals and actions, has three different kinds of precondition that can be used to represent information goals either alone or in combination (Etzioni et al., 1992). As well as `satisfy` preconditions, which may be achieved through actions or through observation, UWL has `hands-off` preconditions indicating that the value of propositions must not be changed in order to achieve the subgoal, and `find-out` preconditions. The latter are in some ways similar to preconditions for `know-if` propositions in Cassandra. A precondition such as `(find-out (P . v))` tells the planner to ascertain whether or not P has truth value v. Under certain circumstances this type of precondition may be achieved by an action that changes the value of P. Knowledge goals may thus be represented by `find-out` preconditions or `satisfy` preconditions (often used in conjunction with `hands-off` preconditions). Etzioni et al. argue that knowledge goals should only be achieved through actions that change the value of the proposition in question when that change is required for another purpose in the plan. We believe that this is an unnecessary limitation, and that in some circumstances enforcement actions may be the best way of achieving knowledge goals.

### 7.3 Probabilistic and Decision-theoretic Planning

When constructing plans, Cassandra recognizes the presence of uncertainty but not its extent. Other planners specifically address issues of probability: for example, BURIDAN constructs plans whose probability of achieving the goal is above a given threshold (Kushmerick et al., 1995); and DRIPS uses the utility of the different possible outcome of various plans to choose the one with the highest expected utility (Haddawy & Suwandi, 1994). Neither BURIDAN nor DRIPS constructs contingency plans, *i.e.*, plans that involve alternative courses of action to be performed in different circumstances. C-BURIDAN, which is based on BURIDAN, constructs contingency plans that are likely to succeed (Draper et al., 1994b, 1994a). It represents an extension of CNLP in the direction of decision-theoretic planning.

Probabilistic planners use information about the probabilities of the possible uncertain outcomes to construct plans that are likely to succeed. Cassandra, on the other hand, cannot use such information and constructs plans that are guaranteed to succeed. Probabilistic planning, because it relies on explicit probabilities, is both more and less powerful than the deterministic contingency planning performed by Cassandra. Cassandra cannot use information about probabilities but it can construct plans in circumstances in which no such information is available. For example, in order to solve the bomb-in-the-toilet problem, c-buridan would have to have some information, or at least make an assumption, about the probabilities of the bomb being in each package. Whatever assumptions are made might turn out to be wrong, thus invalidating the basis of the plan.

We believe that it would be possible to build a probabilistic planner using ideas from both c-buridan and Cassandra. Because of the explicit representation of decisions in Cassandra, such a planner would provide an excellent opportunity for investigating the use of different decision procedures. C-buridan relies on having full knowledge of all the probabilities at the time that it constructs its plans. This knowledge, like any other, may not be available until the plan is executed. It would be relatively simple to add decision procedures to Cassandra's decision representation that depend on information about probabilities, *e.g.*, to follow a particular course of action if the probability of a given outcome exceeds a certain value. The introduction of such decision procedures might, of course, result in the introduction of knowledge goals to determine probabilities, possibly leading eventually to a system that would construct plans to perform empirical studies to determine probabilities.

A problem associated with contingency planning is that of branch merging, *i.e.*, the determination of whether two steps in separate branches can be treated as the same step. C-buridan performs full merging: this is an effect of the probabilistic algorithm on which it is based. Adding this capability to Cassandra is an area of future work. A major problem encountered when considering branch merging is how to identify the variables in the different branches with each other: c-buridan's representations do not include variables, so the problem does not arise. This may cause difficulties in the adaptation of c-buridan's merging mechanism for Cassandra's use.

An advantage of combining probabilistic planning and contingency planning is the resulting ability to judge whether it is worth planning for a given contingency. One of the limitations of Cassandra in its present form is the requirement that every possible contingency be planned for. In complex situations this makes the resulting plans cumbersome. Moreover, Cassandra's performance deteriorates with the number of distinct branches in the plan. The cost of determining that the presence of a particular branch would not significantly change the probability of the plan's success might well be much less than the cost of constructing that branch. This is an interesting issue to be considered in the future.

### 7.4 Interleaving Planning and Execution

Although Cassandra's plans may include sensing actions, with the course of action that will actually be executed depending on the results of those actions, Cassandra does not interleave planning with execution. Plans are fully specified before they are executed. In some circumstances this is clearly very inefficient. Consider, for example, how Cassandra constructs a plan to open a combination safe (see Section 6.4). It requires prior knowledge of all possible combinations, and then constructs a plan with a branch for each combination.

An obvious alternative would be to construct a plan that was fully specified up to the information-gathering step, execute the plan to that stage and, once the information has been gathered, construct the rest of the plan.[10] This could be done in Cassandra by introducing another type of decision procedure, that of planning to achieve the goal, and assuming that it would always be possible to find a plan to achieve the goal. This is a strong assumption, but would certainly be valid in cases such as the problem of opening a safe. This is an area of future work. Interleaving planning and execution in this way would have the advantage that it would not be necessary to plan for contingencies that do not actually arise. It would however lose some of the advantages of planning in advance. For example, possible interference between actions performed before and after the information gathering might be missed, leading the planner to find suboptimal plans. Indeed, as sensing actions may in general change the world, executing them before full construction of a viable plan might have the unfortunate result of making the achievement of the goal impossible.

---

10. See Section 8.2 for further discussion of this issue and an alternative approach.

Planners that interleave planning and execution include IPEM (Ambros-Ingerson & Steel, 1988), XII (Golden, Etzioni, & Weld, 1994) and Sage (Knoblock, 1995). All three use the same basic interleaving technique: only when no further planning is possible are steps executed. They thus do not set out to decide in advance exactly when further planning will be necessary, and their plans do not include explicit provision for further planning. The effects of different interleaving strategies were investigated in the design of BUMP (Olawsky & Gini, 1990). In the Continue Elsewhere strategy as much preplanning as possible was performed; in the Stop and Execute strategy, goals defined in terms of sensor readings were executed as soon as they were encountered. It was found that neither strategy had a clear advantage over the other, in that both strategies sometimes produced plans that were suboptimal or that might fail.

### 7.5 Reactive Planning

A different approach to the problem of planning under uncertainty is taken in the *reactive planning* paradigm. In this approach, no specific sequence of actions is planned in advance. Just as for contingency planning, the planner is given a set of initial conditions and a goal. However, instead of producing a plan with branches, it produces a set of condition-action rules: for example, *universal plans* (Schoppers, 1987) or Situated Control Rules (SCRs) (Drummond, 1989).

In theory, a reactive planning system can handle exogenous events as well as uncertain effects and unknown initial conditions: it is possible to provide a reaction rule for every possible situation that may be encountered, whether or not the circumstances that would lead to it can be envisaged. In contrast, a contingency planner such as Cassandra cannot handle exogenous events as it cannot predict them. Cassandra and other contingency planners focus their planning effort on circumstances that are predicted to be possible (or likely, in the case of a probabilistic contingency planner such as C-BURIDAN).

It would be possible to represent Cassandra's contingency plans as sets of condition-action rules, by using the causal links and preconditions to specify the conditions in which each action should be performed. However, more reasoning is required at execution time to use reaction rules than is required to execute a contingency plan. Instead of simply executing the next step in the plan, reasoning only at branch points, the use of reaction rules requires the evaluation of conditions on every cycle in order to select the relevant rule.

## 8. Discussion

We have described Cassandra, a partial-order contingency planner that can represent uncertain outcomes and construct contingency plans for those outcomes. The design of Cassandra is based on a coherent view of the issues arising in planning under uncertainty. It recognizes that, in an uncertain world, a distinction must be drawn between the actual state of the world and the planner's model of it; it instantiates an intuitively natural account of why knowledge goals exist and how they arise; and it bases its treatment of plan branching on the requirements of the agent that will execute the plan. As a result, Cassandra explicitly plans to gather information and allows information-gathering actions to be fully general. The coherence of its design provides a solid base for more advanced capabilities such as the use of varying decision-making procedures.

## 8.1 Contributions

The principal contribution of this work lies in the explicit representation of decision steps and the implications this has for the handling of knowledge goals. Cassandra is, we believe, the first planner in which decisions are represented as explicit actions in the plans that it constructs. Cassandra's knowledge goals arise specifically from the need to decide between alternative courses of action, as preconditions of the decision actions. Cassandra is thus consistent with the view that planning is the process of making decisions in advance. In this view, contingency plans are plans that defer some decisions until the information on which they are based will be available (Pryor, 1995). Different plan branches correspond to different decision outcomes.

Through its use of explicit decision steps, Cassandra distinguishes between sensing or information-gathering actions on the one hand, and decision making on the other. One important reason for making this distinction is that a decision may depend on more than one piece of information, each available through performing different actions. In addition, separating information-gathering from decision-making provides a basis for introducing alternative methods for making decisions. For example, the extension to Cassandra described in Section 6.5.5 introduces a type of decision that directs the executing agent to perform all branches resulting from a given source of uncertainty, which allows the construction of plans that can succeed in situations in which there is no way of telling what the actual outcome is (*e.g.*, the bomb-in-the-toilet problem). We believe that the explicit representation of different methods for making decisions is an important direction for future research.

Because knowledge goals arise as preconditions of decisions in Cassandra, the need to know whether a particular plan branch will work is distinguished from the need to know the actual outcome of an uncertainty. Cassandra does not plan to determine outcomes unless they are relevant to the achievement or otherwise of its goals. Moreover, Cassandra does not treat knowledge goals as special cases: plans to achieve them may be as complex as plans to achieve any other goals. As well as planning to achieve knowledge goals that arise as preconditions of decisions, Cassandra can also produce plans for top-level knowledge goals.

Two other features of Cassandra are worth noting: the flexibility afforded by its labeling scheme; and the potential for learning and adaptation afforded by its representation of uncertainty.

Cassandra's labeling scheme, although complex, allows the agent executing the plan to distinguish between three classes of action: those that must be executed in a given contingency; those that must not; and those whose execution will not affect the achievement of the goal in that contingency.[11] This feature paves the way for the extension described above that allows Cassandra to build plans requiring the execution of all branches resulting from a source of uncertainty.

Cassandra's representation makes no assumptions as to the intrinsic nature of uncertainty. An unknown precondition simply denotes that the information as to what context will produce a particular effect from an action is not available to the planner. It may be that this information is in principle unknowable (in domains involving quantum effects, for example); it is much more likely that the uncertainty results from the limitations of the planner or of the information available to it. In general, an agent operating in a real-world domain will be much more effective if it can learn to improve its performance and adapt to changing conditions. The use of unknown preconditions to represent uncertainty means that in some circumstances it would be relatively simple to incorporate the results of such learning and adaptation into the planner's domain knowledge. For example, the planner might discover how to predict certain outcomes: it could then change the unknown preconditions into ones reflecting the new knowledge. If, on the other hand, it discovered that predicted effects were consistently failing to occur, it could change the relevant preconditions into unknown ones.

## 8.2 Limitations

Cassandra is one of an increasing number of planners that aim to extend the techniques of classical planning to more realistic domains. Cassandra is designed to operate in domains in which two of the three principal constraints observed by classical planners are relaxed: namely, we allow non-deterministic actions and incomplete knowledge of the initial conditions. Cassandra is, however, subject to the third constraint, that changes do not take place except as a result of actions specified in the plan. This clearly limits its effectiveness in many real-world domains. Moreover, there are limits on the extent of the nondeterminism and incompleteness of knowledge that are handled. Cassandra's plans will not necessarily achieve their goals if sources of uncertainty are ignored, or if all possible outcomes are not specified.

Cassandra cannot make use of information about how likely particular outcomes are, unlike probabilistic or decision-theoretic planners; it cannot plan to interleave planning and execution; and it does not provide reaction rules for all possible circumstances. It can only solve problems for which there are valid plans involving ways of discriminating between possible outcomes; the algorithm given here cannot solve the original version of the bomb-in-the-toilet problem, although the extension described in Section 6.5.5 can do so (Collins & Pryor, 1995).

The algorithm described in this paper has two major practical limitations: first, the plans it produces are often more complex than necessary; and second, the time taken to produce plans precludes its use on all except simple problems.

The complexity of Cassandra's plans results from the necessity of planning for every contingency and from the lack of branch merging. For example, suppose you had to open a combination safe so that you could obtain the money to pay for an evening out. Cassandra's plan for the goal of enjoying an evening out would have one branch for each possible safe combination. Each branch would start off with the actions to open the safe, which are different for each combination, and would continue with the actions of going to a restaurant and then to the movies, say, which would be identical in each branch. A simpler plan would merge the separate branches after the safe had been opened. The consideration of methods for branch merging is an area of future work (see Sections 6.5.4 and 7.3).

---

11. Not all agents can make use of this information, as there is no guarantee that the third type of step will actually be executable.

In some circumstances, such as in this example, plan complexity could be reduced through the use of run-time variables, which were introduced in IPEM (Ambros-Ingerson & Steel, 1988) and used in SENSp (Etzioni et al., 1992) (see Section 7). When the only uncertainty is in the value that an action parameter takes (which is the case when opening a combination safe) it would be possible to use a run-time variable to represent that parameter, obviating the need for separate plan branches. Implementing this strategy would require effective methods for determining when the effects of uncertainty are limited to parameter values. In general, this notion indicates a possible approach to the problem of branch merging: that of taking a least commitment approach to variable binding, in the same way that a least commitment approach is taken to step ordering in a partial order planner. This would then allow the concept of "conditional" variable binding: a variable binding could be labeled as being required or forbidden in a given contingency.

We have not analyzed the complexity of Cassandra's algorithm, but we believe it to be exponential. This is because of the effect of multiple plan branches, whose presence not only increases the number of steps in a plan but also increases the number of potential interactions and the number of ways of resolving them. Certainly, our subjective impression is that Cassandra runs even more slowly than other planners in the SNLP family. Effective domain-independent search control heuristics are difficult to find, and in many of the (toy) domains in which we have used Cassandra even problem-specific heuristics are hard to come by.

### 8.3 Conclusion

Cassandra is a planning system based firmly in the classical planning paradigm. Many of its strengths and weaknesses are those of other classical planning systems. For example, we believe that under certain circumstances its plans will be valid and that it is guaranteed to find a valid plan if one exists. However, the techniques it uses are valid only in limited circumstances, and its computational complexity is such as to make direct scaling up unlikely to be feasible.

In our view, the principal strengths of Cassandra arise from the explicit representation of decisions in its plans. We have shown how this use of decisions provides a natural account of how knowledge goals arise during the planning process. We have also sketched how decisions can be used as the basis of extensions that provide added functionality. A new type of decision allows fail-safe plans, which can provide a method of solving problems such as the bomb-in-the-toilet problem (Section 6.5.5); and another type of decision may provide an effective method of interleaving planning and execution (Section 7.4).

We believe that the use of explicit decision procedures will enable the extension of the range of applicability of techniques of classical planning. In general, the idea of constructing a single plan that will succeed in all circumstances is, we feel, unlikely to be productive:

PRYOR & COLLINS

the real world is complex and uncertain enough that trying to predict its behavior in detail is simply impossible. However, the use of decision procedures that, for example, involve probabilistic techniques or interleave planning and execution, appears likely to provide a flexible framework that, although inevitably sacrificing completeness and correctness, will provide a basis for effective, practical planning in the real world.

## Appendix A. Cassandra's Plans

This appendix shows the plans constructed by Cassandra for the examples in the body of the paper. Each plan consists of initial conditions, plan steps and goals. The initial conditions are shown at the top of the plan. Those that are unknown are shown as depending on a particular contingency. The plan steps are shown next. Each is shown with a number denoting its order in the plan. The numbers in parentheses show the order in which the steps were added to the plan. To the right of each step are its contingency labels. For brevity, the individual effects of each step are always omitted and the links that establish the step's enabling and secondary preconditions are often omitted.

Finally, at the bottom of the plan come the goal conditions. The goal is stated first, then each contingency goal is shown with the links that establish it. As usual, contingency labels are to the right.

### A.1 A Plan to Get to Evanston

This is the plan shown in Figure 3 and discussed in Section 2.3.2. Note the decision-step with a single active decision-rule. This is the situation discussed in the comments on one-sided contingencies in Section 6.5: the route using Western is quicker when it is clear, while the Ashland route is slower but always possible.

```
Initial:      When [TRAFFICOS: GOOD] (NOT (TRAFFIC-BAD))
              When [TRAFFICOS: BAD] (TRAFFIC-BAD)
              (AND (AT START) (ROAD WESTERN) (ROAD BELMONT) (ROAD ASHLAND))

Step  1 (4): (GO-TO-WESTERN-AT-BELMONT)              YES: [TRAFFICOS: GOOD BAD]
              (AND (NOT (AT START)) (ON WESTERN) (ON BELMONT))
                 0 -> (AT START)

Step  2 (3): (CHECK-TRAFFIC-ON-WESTERN)
              (KNOW-IF (TRAFFIC-BAD))
                 1 -> (ON WESTERN)

Step  3 (2): (DECIDE TRAFFICOS)
              (and (NOT (TRAFFIC-BAD))
                    T                  ) => [TRAFFICOS: GOOD]
              (and T                   ) => [TRAFFICOS: BAD]
                 2 -> (KNOW-IF (TRAFFIC-BAD))

Step  4 (6): (TAKE-BELMONT)                          YES: [TRAFFICOS: BAD]
                                                     NO : [TRAFFICOS: GOOD]
              (AND (NOT (ON WESTERN)) (ON ASHLAND))
                 1 -> (ON BELMONT)
```

```
Step  5 (5): (TAKE-ASHLAND)                        YES: [TRAFFICOS: BAD]
                                                   NO : [TRAFFICOS: GOOD]
             (AT EVANSTON)
                 4 -> (ON ASHLAND)                 NO : [TRAFFICOS: GOOD]

Step  6 (1): (TAKE-WESTERN)                        YES: [TRAFFICOS: GOOD]
                                                   NO : [TRAFFICOS: BAD]
             (AT EVANSTON)
                 1 -> (ON WESTERN)                 NO : [TRAFFICOS: BAD]
                 0 -> (NOT (TRAFFIC-BAD))          NO : [TRAFFICOS: BAD]

Goal:        (AT EVANSTON)

             GOAL                                  YES: [TRAFFICOS: BAD]
                 5 -> (AT EVANSTON)                NO : [TRAFFICOS: GOOD]

             GOAL                                  YES: [TRAFFICOS: GOOD]
                 6 -> (AT EVANSTON)                NO : [TRAFFICOS: BAD]

Complete!
```

### A.2 Disarming a Bomb

This is the plan shown in Figures 6 and 7 and discussed in Section 4.1.1. Note that both moving steps and both dunking steps are always possible, but each is only necessary in one outcome of the uncertainty. A fail-safe plan (see Section 6.2) is therefore possible.

```
Initial:     When [UNKOS: 02] (CONTAINS PACKAGE-2 BOMB)
             When [UNKOS: 01] (CONTAINS PACKAGE-1 BOMB)
             (AND (AT PACKAGE-1 RUG) (AT PACKAGE-2 RUG))

Step  1 (5): (X-RAY PACKAGE-2)
             (KNOW-IF (CONTAINS PACKAGE-2 BOMB))

Step  2 (3): (X-RAY PACKAGE-1)
             (KNOW-IF (CONTAINS PACKAGE-1 BOMB))

Step  3 (2): (DECIDE UNKOS)
             (and (CONTAINS PACKAGE-2 BOMB)
                  T                 ) => [UNKOS: 02]
             (and (CONTAINS PACKAGE-1 BOMB)
                  T                 ) => [UNKOS: 01]
                 1 -> (KNOW-IF (CONTAINS PACKAGE-2 BOMB))
                 2 -> (KNOW-IF (CONTAINS PACKAGE-1 BOMB))

Step  4 (7): (MOVE RUG TOILET PACKAGE-1)           YES: [UNKOS: 01]
             (AND (NOT (AT PACKAGE-1 RUG)) (AT PACKAGE-1 TOILET))
                 0 -> (AT PACKAGE-1 RUG)

Step  5 (6): (MOVE RUG TOILET PACKAGE-2)           YES: [UNKOS: 02]
             (AND (NOT (AT PACKAGE-2 RUG)) (AT PACKAGE-2 TOILET))
                 0 -> (AT PACKAGE-2 RUG)

Step  6 (4): (DUNK PACKAGE-2)                      YES: [UNKOS: 02]
             (WET PACKAGE-2)
```

```
                      5 -> (AT PACKAGE-2 TOILET)
                  (DISARMED BOMB)
                      0 -> (CONTAINS PACKAGE-2 BOMB)  NO : [UNKOS: 01]

Step  7 (1): (DUNK PACKAGE-1)                         YES: [UNKOS: 01]
                  (WET PACKAGE-1)
                      4 -> (AT PACKAGE-1 TOILET)
                  (DISARMED BOMB)
                      0 -> (CONTAINS PACKAGE-1 BOMB)  NO : [UNKOS: 02]

Goal:        (DISARMED BOMB)

                  GOAL                                YES: [UNKOS: 02]
                      6 -> (DISARMED BOMB)            NO : [UNKOS: 01]

                  GOAL                                YES: [UNKOS: 01]
                      7 -> (DISARMED BOMB)            NO : [UNKOS: 02]

Complete!
```

### A.3 Fetching a Package

The plan in Figure 8, discussed in Section 4.1.3, involves just one source of uncertainty and hence contains just one decision-step. There are two possible ways of achieving the goal, one for each outcome of the uncertainty.

```
Initial:     (AVAILABLE CAR-1)
             When [LOCOS: B] (PACKAGE-AT LOCATION-2)
             When [LOCOS: A] (PACKAGE-AT LOCATION-1)
             (AND (IS-CAR CAR-1) (IS-CAR CAR-2) (LOCATION LOCATION-1)
                  (LOCATION LOCATION-2))

Step  1 (2): (ASK-ABOUT-PACKAGE)
                  (KNOW-IF (PACKAGE-AT LOCATION-2))
                      0 -> (LOCATION LOCATION-2)
                  (KNOW-IF (PACKAGE-AT LOCATION-1))
                      0 -> (LOCATION LOCATION-1)

Step  2 (1): (DECIDE LOCOS)
                  (and (PACKAGE-AT LOCATION-2)
                       T                  ) => [LOCOS: B]
                  (and (PACKAGE-AT LOCATION-1)
                       T                  ) => [LOCOS: A]
                      1 -> (KNOW-IF (PACKAGE-AT LOCATION-2))
                      1 -> (KNOW-IF (PACKAGE-AT LOCATION-1))

Step  3 (4): (DRIVE CAR-1 LOCATION-1)                 YES: [LOCOS: A]
                  (AT LOCATION-1)
                      0 -> (AVAILABLE CAR-1)

Step  4 (3): (DRIVE CAR-1 LOCATION-2)                 YES: [LOCOS: B]
                  (AT LOCATION-2)
                      0 -> (AVAILABLE CAR-1)

Goal:        (AND (AT ?LOC) (PACKAGE-AT ?LOC))
```

```
            GOAL                                 YES: [LOCOS: B]
                4 -> (AT LOCATION-2)
                0 -> (PACKAGE-AT LOCATION-2)     NO : [LOCOS: A]

            GOAL                                 YES: [LOCOS: A]
                3 -> (AT LOCATION-1)
                0 -> (PACKAGE-AT LOCATION-1)     NO : [LOCOS: B]
```

Complete!

### A.4 Fetching Another Package

The plan in Figure 9, discussed in Section 4.1.3, has two sources of uncertainty and two decision-steps. There are four possible ways of achieving the goal, one for each combination of the outcomes of the two sources of uncertainty.

```
Initial:        When [CAROS: C2] (AVAILABLE CAR-2)
                When [CAROS: C1] (AVAILABLE CAR-1)
                When [LOCOS: B] (PACKAGE-AT LOCATION-2)
                When [LOCOS: A] (PACKAGE-AT LOCATION-1)
                (AND (IS-CAR CAR-1) (IS-CAR CAR-2) (LOCATION LOCATION-1)
                     (LOCATION LOCATION-2))

Step  1 (5): (ASK-ABOUT-CAR)                     YES: [LOCOS: A B]

                (KNOW-IF (AVAILABLE CAR-2))
                   0 -> (IS-CAR CAR-2)
                (KNOW-IF (AVAILABLE CAR-1))
                   0 -> (IS-CAR CAR-1)

Step  2 (4): (DECIDE CAROS)                      YES: [LOCOS: A B]
                (and (AVAILABLE CAR-2)
                     T              ) => [CAROS: C2]
                (and (AVAILABLE CAR-1)
                     T              ) => [CAROS: C1]
                   1 -> (KNOW-IF (AVAILABLE CAR-2))
                   1 -> (KNOW-IF (AVAILABLE CAR-1))

Step  3 (2): (ASK-ABOUT-PACKAGE)                 YES: [CAROS: C2 C1]

                (KNOW-IF (PACKAGE-AT LOCATION-2))
                   0 -> (LOCATION LOCATION-2)
                (KNOW-IF (PACKAGE-AT LOCATION-1))
                   0 -> (LOCATION LOCATION-1)

Step  4 (1): (DECIDE LOCOS)                      YES: [CAROS: C2 C1]
                (and (PACKAGE-AT LOCATION-2)
                     T              ) => [LOCOS: B]
                (and (PACKAGE-AT LOCATION-1)
                     T              ) => [LOCOS: A]
                   3 -> (KNOW-IF (PACKAGE-AT LOCATION-2))
                   3 -> (KNOW-IF (PACKAGE-AT LOCATION-1))

Step  5 (8): (DRIVE CAR-2 LOCATION-1)            YES: [LOCOS: A][CAROS: C2]
```

```
                                                    NO  : [CAROS: C1]
                   (AT LOCATION-1)
                      0 -> (AVAILABLE CAR-2)        NO  : [CAROS: C1]

Step  6 (6): (DRIVE CAR-2 LOCATION-2)               YES : [LOCOS: B][CAROS: C2]
                                                    NO  : [CAROS: C1]
                   (AT LOCATION-2)
                      0 -> (AVAILABLE CAR-2)        NO  : [CAROS: C1]

Step  7 (7): (DRIVE CAR-1 LOCATION-1)               YES : [LOCOS: A][CAROS: C1]
                                                    NO  : [CAROS: C2]
                   (AT LOCATION-1)
                      0 -> (AVAILABLE CAR-1)        NO  : [CAROS: C2]

Step  8 (3): (DRIVE CAR-1 LOCATION-2)               YES : [LOCOS: B][CAROS: C1]
                                                    NO  : [CAROS: C2]
                   (AT LOCATION-2)
                      0 -> (AVAILABLE CAR-1)        NO  : [CAROS: C2]

Goal:        (AND (AT ?LOC) (PACKAGE-AT ?LOC))

                   GOAL                             YES : [LOCOS: A][CAROS: C2]
                      5 -> (AT LOCATION-1)          NO  : [CAROS: C1]
                      0 -> (PACKAGE-AT LOCATION-1)  NO  : [LOCOS: B]

                   GOAL                             YES : [LOCOS: B][CAROS: C2]
                      6 -> (AT LOCATION-2)          NO  : [CAROS: C1]
                      0 -> (PACKAGE-AT LOCATION-2)  NO  : [LOCOS: A]

                   GOAL                             YES : [LOCOS: B][CAROS: C1]
                      8 -> (AT LOCATION-2)          NO  : [CAROS: C2]
                      0 -> (PACKAGE-AT LOCATION-2)  NO  : [LOCOS: A]

                   GOAL                             YES : [LOCOS: A][CAROS: C1]
                      7 -> (AT LOCATION-1)          NO  : [CAROS: C2]
                      0 -> (PACKAGE-AT LOCATION-1)  NO  : [LOCOS: B]

Complete!
```

### A.5 Tossing a Coin

In Section 4.2.3 we described a plan for ending up with a flat coin. The decision in this plan does not distinguish between the coin landing heads-up and tails-up—the decision rules are ambiguous.

```
Initial:     (HOLDING-COIN)

Step  1 (2): (TOSS-COIN)
             (AND (NOT (HOLDING-COIN)) (ON-TABLE))
                0 -> (HOLDING-COIN)

Step  2 (4): (INSPECT-COIN)
             (AND (KNOW-IF (FLAT-COIN)) (KNOW-IF (HEADS-UP))
               (KNOW-IF (TAILS-UP)) (KNOW-IF (ON-EDGE)))
```

```
Step  3 (3): (DECIDE UNK2S)
             (and (FLAT-COIN)
                  T                    ) => [UNK2S: H]
             (and (FLAT-COIN)
                  T                    ) => [UNK2S: T]
             (and (ON-EDGE)
                  T                    ) => [UNK2S: E]
                2 -> (KNOW-IF (FLAT-COIN))
                2 -> (KNOW-IF (ON-EDGE))

Step  4 (1): (TIP-COIN)                     YES: [UNK2S: E]
                                            NO : [UNK2S: H T]
             (FLAT-COIN)
                1 -> (ON-EDGE)              NO : [UNK2S: H T]

Goal:        (FLAT-COIN)

             GOAL                           YES: [UNK2S: T]
                1 -> (FLAT-COIN)            NO : [UNK2S: H E]

             GOAL                           YES: [UNK2S: H]
                1 -> (FLAT-COIN)            NO : [UNK2S: T E]

             GOAL                           YES: [UNK2S: E]
                4 -> (FLAT-COIN)            NO : [UNK2S: H T]

Complete!
```

## A.6 Tossing Another Coin

The plan in Figure 11 has two decisions with unambiguous decision-rules. There are four ways of achieving the goal in this plan, because there are two sources of uncertainty.

```
Initial:     (HOLDING-COIN)

Step  1 (1): (TOSS-COIN)
             (AND (NOT (HOLDING-COIN)) (ON-TABLE) (KNOW-IF (FLAT-COIN))
              (KNOW-IF (HEADS-UP)) (KNOW-IF (TAILS-UP)) (KNOW-IF (ON-EDGE)))
                0 -> (HOLDING-COIN)

Step  2 (2): (DECIDE TOSS1S)
             (and (FLAT-COIN)
                  (HEADS-UP)
                  T                    ) => [TOSS1S: H]
             (and (ON-EDGE)
                  T                    ) => [TOSS1S: E]
             (and (FLAT-COIN)
                  (TAILS-UP)
                  T                    ) => [TOSS1S: T]
                1 -> (KNOW-IF (ON-EDGE))
                1 -> (KNOW-IF (FLAT-COIN))
                1 -> (KNOW-IF (TAILS-UP))
                1 -> (KNOW-IF (HEADS-UP))

Step  3 (4): (TIP-COIN)                     YES: [TOSS1S: E]
```

```
                                                    NO : [TOSS1S: T H]
                (AND (FLAT-COIN) (KNOW-IF (HEADS-UP)) (KNOW-IF (TAILS-UP)))
                    1 -> (ON-EDGE)               NO : [TOSS1S: H T]

Step  4 (5): (DECIDE TIP4S)                      YES: [TOSS1S: E]
                                                 NO : [TOSS1S: T H]
                (and (TAILS-UP)
                     T                ) => [TIP4S: T]
                (and (HEADS-UP)
                     T                ) => [TIP4S: H]
                    3 -> (KNOW-IF (TAILS-UP))    NO : [TOSS1S: T H]
                    3 -> (KNOW-IF (HEADS-UP))    NO : [TOSS1S: T H]

Step  5 (3): (TURN-OVER)                         YES: [TOSS1S: T]
                                                 NO : [TOSS1S: E H]
                    1 -> (FLAT-COIN)             NO : [TOSS1S: H E]

                (HEADS-UP)
                    1 -> (TAILS-UP)              NO : [TOSS1S: H E]

Step  6 (6): (TURN-OVER)                         YES: [TOSS1S: E][TIP4S: T]
                                                 NO : [TOSS1S: T H][TIP4S: H]
                    3 -> (FLAT-COIN)             NO : [TOSS1S: T H]

                (HEADS-UP)
                    3 -> (TAILS-UP)              NO : [TOSS1S: T H][TIP4S: H]

Goal:       (AND (FLAT-COIN) (HEADS-UP))

                GOAL                             YES: [TOSS1S: E][TIP4S: T]
                    3 -> (FLAT-COIN)             NO : [TOSS1S: T H]
                    6 -> (HEADS-UP)              NO : [TOSS1S: T H][TIP4S: H]

                GOAL                             YES: [TOSS1S: E][TIP4S: H]
                    3 -> (FLAT-COIN)             NO : [TOSS1S: T H]
                    3 -> (HEADS-UP)              NO : [TOSS1S: H T][TIP4S: T]

                GOAL                             YES: [TOSS1S: T]
                    1 -> (FLAT-COIN)             NO : [TOSS1S: H E]
                    5 -> (HEADS-UP)              NO : [TOSS1S: E H]

                GOAL                             YES: [TOSS1S: H]
                    1 -> (FLAT-COIN)             NO : [TOSS1S: T E]
                    1 -> (HEADS-UP)              NO : [TOSS1S: T E]

Complete!
```

## A.7 Opening a Door

In Section 4.2.4 we described a plan for opening a locked door without a key; it is depicted in Figure 12. The plan that Cassandra produces for this situation is shown here. Even though no preconditions of the pick step depend on any effect of the kick step, the former cannot be performed if the lock is broken as a result of kicking the door. The decision-rules reflect this dependence.

```
Initial:       (LOCK-INTACT)

Step  1 (2): (KICK)

Step  2 (4): (LOOK)
             (AND (KNOW-IF (LOCKED)) (KNOW-IF (LOCK-INTACT))
              (KNOW-IF (FOOT-BROKEN)))

Step  3 (3): (DECIDE KICK2S)
             (and ((LOCK-INTACT))
                  T                 ) => [KICK2S: F]
             (and (NOT (LOCKED))
                  T                 ) => [KICK2S: L]
                 2 -> (KNOW-IF (LOCKED))

Step  4 (6): (PICK)                            YES: [KICK2S: F]
                                               NO : [KICK2S: L]
               (NOT (LOCKED))
                 0 -> (LOCK-INTACT)            NO : [KICK2S: L]

Step  5 (5): (OPEN-DOOR)                       YES: [KICK2S: F]
                                               NO : [KICK2S: L]
               (OPEN)
                 4 -> (NOT (LOCKED))           NO : [KICK2S: L]

Step  6 (1): (OPEN-DOOR)                       YES: [KICK2S: L]
                                               NO : [KICK2S: F]
               (OPEN)
                 1 -> (NOT (LOCKED))           NO : [KICK2S: F]

Goal:          (OPEN)

               GOAL                            YES: [KICK2S: F]
                 5 -> (OPEN)                   NO : [KICK2S: L]

               GOAL                            YES: [KICK2S: L]
                 6 -> (OPEN)                   NO : [KICK2S: F]

Complete!
```

## Acknowledgements

Thanks to Dan Weld and Tony Barrett for supplying the UCPOP code, Mark Peot and Robert Goldman for their comments on earlier drafts, Will Fitzgerald for many useful discussions, and the anonymous reviewers for their constructive and helpful criticism. Much of this work was performed while the first author was a student at the Institute for the Learning Sciences, Northwestern University. This work was supported in part by the AFOSR under grant number AFOSR-91-0341-DEF. The Institute for the Learning Sciences was established in 1989 with the support of Andersen Consulting, part of The Arthur Andersen Worldwide Organization. The Institute receives additional support from Ameritech and North West Water, Institute Partners, and from IBM.